\documentclass[11pt]{article}

\usepackage[final]{acl}

\usepackage{times}
\usepackage{latexsym}

\usepackage[T1]{fontenc}

\usepackage[utf8]{inputenc}

\usepackage{microtype}

\usepackage{inconsolata}

\usepackage{graphicx}
\usepackage{enumitem}

\usepackage[linesnumbered,ruled,vlined]{algorithm2e}
\usepackage[table]{xcolor}       % colors
\usepackage{enumitem}
\usepackage{amsmath}
\usepackage{tabularx}
\usepackage{boldline}
\usepackage{booktabs}
\usepackage{arydshln}
\usepackage{multirow}
\usepackage{colortbl}
\usepackage{tabularx}
\usepackage{makecell}
\usepackage{listings}

\title{From Reasoning to Generalization:
Knowledge-Augmented LLMs for the ARC Benchmark}

\author{Chao Lei, Krista A. Ehinger, Nir Lipovetzky, Yanchuan Chang \\
    School of Computing and Information Systems, The University of Melbourne, Australia\\
\texttt{clei1@student.unimelb.edu.au,
}
\\
\texttt{\{kris.ehinger, nir.lipovetzky, yanchuan.chang\}@unimelb.edu.au}}

\begin{document}
\maketitle
\begin{abstract}

Despite extensive research on reasoning-oriented LLMs, core cognitive faculties of human intelligence, such as abstract reasoning and generalization, remain underexplored. To address this, we evaluate recent reasoning-oriented LLMs on the Abstraction and Reasoning Corpus (ARC) benchmark, which explicitly demands both faculties. We formulate ARC as a program synthesis task and propose nine candidate solvers. Experimental results show that repeated-sampling planning-aided code generation (RSPC) achieves the highest test accuracy and demonstrates consistent generalization across most LLMs. To further improve performance, we introduce Knowledge Augmentation for Abstract Reasoning (KAAR), which encodes \textit{core knowledge} priors within an ontology that classifies priors into three hierarchical levels based on their dependencies. KAAR progressively expands LLM reasoning capacity by gradually augmenting priors at each level, and invokes RSPC to generate candidate solutions after each augmentation stage. 
Empirical results show that KAAR maintains strong generalization and consistently outperforms non-augmented RSPC  across all evaluated LLMs, achieving around 5\% absolute gains and up to 64.52\% relative improvement.  Our code is available at \url{https://github.com/you68681/kaar}. %Nonetheless, ARC remains a challenging benchmark for reasoning-oriented LLMs, highlighting future avenues of progress in LLMs.
\end{abstract}

\section{Introduction}

\begin{figure}[t]
\centering

 \includegraphics[width=0.47\textwidth]{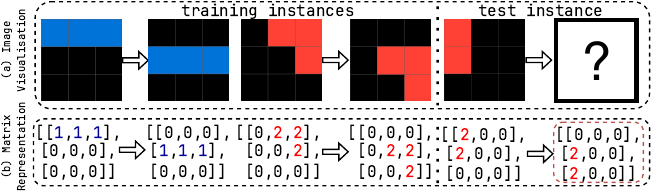}
 
\caption{Example ARC problem (\textit{25ff71a9}) with image visualizations (a), including two input-output pairs in the training instances and one input image in the test instance, along with their 2D matrix representations (b). The ground-truth test output is enclosed in a red box. }

\label{fig1}
\end{figure}

Abstract reasoning and generalization are fundamental to understanding and evaluating machine intelligence \citep{malkinski2022review,barrett2018measuring,moskvichev2023conceptarc}. \citet{chollet2019measure} introduced the Abstraction and Reasoning Corpus (ARC) to assess these capabilities of AI systems. In each ARC task, the solver is required to infer generalized rules or procedures from a small set of training instances, typically fewer than five input-output image pairs, and apply them to generate  output images for given input images in test instances (Figure~\ref{fig1} (a)). Each image in ARC is a pixel grid represented as a 2D matrix, where each value denotes a pixel color (Figure~\ref{fig1} (b)). ARC evaluates \textit{broad generalization}, encompassing reasoning over individual input-output pairs and inferring generalized solutions via high-level abstraction, akin to inductive reasoning \cite{peirce1868questions}.

ARC is grounded in \textit{core knowledge} priors that serve as foundational cognitive faculties of human intelligence \cite{spelke2007core}. We note that all ARC solvers are assumed to possess these priors, enabling equitable comparisons between AI systems and human cognitive abilities \cite{chollet2019measure}. An abstract and expressive representation of core knowledge priors is fundamental in ARC to enable human-like reasoning \cite{chollet2019measure}. Specifically, these priors include: (1) \textit{objectness} – aggregating elements into coherent, persistent objects; (2) \textit{geometry and topology} – recognizing and manipulating shapes, symmetries, spatial transformations, and structural patterns (e.g., containment, repetition, projection); (3) \textit{numbers and counting} – counting, sorting, comparing quantities, performing basic arithmetic, and identifying numerical patterns; and (4) \textit{goal-directedness} – inferring purposeful transformations between initial and final states without explicit temporal cues.

\citet{chollet2019measure} suggested approaching ARC tasks as instances of program synthesis. Following this proposal, previous solvers, Abstract Reasoning with Graph Abstractions (ARGA) \citep{xu2022graphs} and Generalized Planning for Abstract Reasoning (GPAR) \citep{lei2024generalized}, a state-of-the-art object-centric solver,  have successfully solved subsets of ARC tasks by searching for program solutions within object-centric domain-specific languages (DSLs). GPAR and KAAR achieve notable generalization benefits from their lifted core knowledge priors representations in DSLs. However, the expansive search space limits their scalability. Recently, reasoning-oriented LLMs trained via reinforcement learning to integrate chain-of-thought reasoning \citep{wei2022chain} further advance program synthesis performance.
Common approaches using LLMs for code generation include repeated sampling, where multiple candidate programs are generated \citep{chen2021evaluating}, followed by best-program selection strategies \citep{li2022competition, chen2022codet, zhang2023coder, ni2023lever}, and code refinement, where initial LLM-generated code is iteratively improved using error feedback from execution results \citep{zhong2024ldb, lei2024planning} or LLM-generated explanations \citep{zhong2024ldb, chen2023teaching, lei2024planning}. We note that ARC presents greater challenges than existing program synthesis benchmarks such as HumanEval \citep{chen2021evaluating}, MBPP \citep{austin2021program}, and LiveCode \citep{jain2024livecodebench}, due to its  emphasis on generalization and abstract reasoning grounded in core knowledge priors, which remain underexplored. This gap motivates our evaluation of reasoning-oriented LLMs on the ARC benchmark and our proposed knowledge augmentation approach to improve their performance. See Appendix~\ref{related_work} for detailed related work.

We begin by systematically assessing  how reasoning-oriented LLMs approach ARC tasks within the program synthesis framework, given solely 2D matrices as input. In total, nine ARC solvers are considered with accuracy on test instances reported as the primary metric. When evaluated on the ARC public evaluation set (400 problems), repeated-sampling planning-aided code generation (RSPC) demonstrates consistent generalization and achieves the highest test accuracy across most evaluated LLMs. We treat the most competitive ARC solver, RSPC, as the solver backbone. We further propose \underline{K}nowledge \underline{A}ugmentation for \underline{A}bstract \underline{R}easoning (KAAR) for solving ARC tasks. 
KAAR refines the core knowledge priors encoded in GPAR and formalizes them through a lightweight ontology that organizes priors into hierarchical levels based on their dependencies. It incrementally augments LLMs with priors at each level through in-context learning. After completing augmentation at each level, KAAR applies the ARC solver backbone (RSPC) to generate the solution. This progressive augmentation enables LLMs to gradually expand their reasoning capabilities and facilitates stage-wise reasoning, aligning with human cognitive development \citep{babakr2019piaget}. Empirical results show that KAAR consistently outperforms RSPC in test accuracy across all evaluated LLMs, while preserving strong generalization. We outline our contributions as follows:

\begin{itemize}[noitemsep, topsep=0pt, leftmargin=*]

    \item We evaluate the abstract reasoning and generalization capabilities of reasoning-oriented LLMs on ARC using nine program synthesis solvers.

    \item  We introduce KAAR, a knowledge augmentation approach for solving ARC problems with LLMs, further improving performance.

    \item  We conduct a comprehensive performance analysis of the evaluated LLMs and proposed ARC solvers, highlighting failure cases and remaining challenges in ARC.

\end{itemize}

\section{Problem Formulation}

We formulate each ARC task as a tuple $\mathcal{P}=\langle I_r, I_t \rangle$, where $I_r$ and $I_t$ are sets of training and test instances. Each instance is an input-output image pair $(i^i, i^o)$, represented as 2D matrices. The goal is to leverage the LLM  $\mathcal{M}$ to generate a solution $s$ based on training instances $I_r$ and test input images $\{i^i \ | \ (i^i,i^o) \in I_t\} $, where $s$ maps each test input $i^i$ to its output $i^o$, i.e., $s(i^i)=i^o$, for $(i^i, i^o) \in I_t$. We note that test input images are visible during the generation of solution $s$, whereas test output images become accessible only after $s$ is produced for validation.  We encode the solution $s$ in different forms, as a text-based plan $p$, or as Python code $c$, optionally guided by $p$. We denote each ARC problem description, comprising $I_r$ and $\{i^i \ | \ (i^i,i^o) \in I_t\}$, as $Q$.

\begin{figure*}[t]
 
\centering
\includegraphics[scale=0.78]{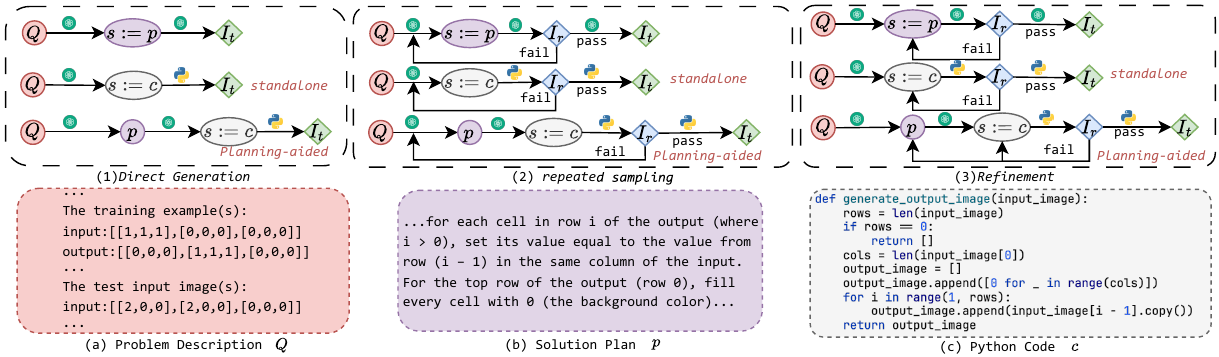}

\caption{ARC solution generation approaches: (1) \textit{direct generation}, (2) \textit{repeated sampling}, and (3) \textit{refinement}, with the GPT-o3-mini input and response fragments (a–c) for solving task \textit{25ff71a9} (Figure~\ref{fig1}). For each approach, when the solution $s$ is code, $s := c$, a plan $p$ is either generated from the problem description $Q$ to guide code generation (\textit{planning-aided}) or omitted (\textit{standalone}). Otherwise, when $s := p$, the plan $p$  serves as the final solution.}

\label{fig2}
\end{figure*}

\section{ARC Solver Backbone}\label{sec:solver_backbone}

{LLMs have shown promise in solving tasks that rely on ARC-relevant priors \citep{deng2024can,meng-etal-2024-llm,ahn-etal-2024-large,zang2025contextual}, suggesting an implicit encoding of core knowledge priors for solving ARC tasks. Unlike previous work that treats LLMs as output image generators \cite{min2023approach,xu2024llms,lee2024reasoning}, we cast each ARC task as a program synthesis problem, which involves generating a solution $s$ from a problem description $Q$, where knowledge priors emerge from LLMs' perception of grid dynamics.}
We consider established LLM-based code generation approaches \citep{zhong2024ldb,lei2024planning,chen2023teaching,islam-etal-2024-mapcoder} as candidate ARC solution generation strategies, illustrated at the top of Figure~\ref{fig2}. 
These include: (1) \textit{direct generation}, where LLM produces the solution $s$ in a single attempt, and then evaluates it on test instances $I_t$; (2) \textit{repeated sampling}, where LLM samples solutions until one passes training instances $I_r$, and then evaluates it on $I_t$; and (3) \textit{refinement}, where LLM iteratively refines an initial solution $s$ based on failures on $I_r$ until it succeeds, followed by evaluation on $I_t$. In addition, we extend the solution representation beyond code to include text-based solution plans. Given the problem description $Q$ as input (Figure~\ref{fig2}, block (a)), all strategies prompt the LLM to generate a solution $s$, represented either as a natural language plan $p$ (block (b)), $s := p$, or as a Python code $c$ (block (c)), $s := c$. For $s:=p$, the solution is derived directly from $Q$. For $s := c$, we explore two modalities: the LLM either generates $c$ directly from $Q$ (\textit{standalone}), or first generates a plan $p$ for $Q$, which is then concatenated with $Q$ to guide subsequent code development (\textit{planning-aided}), a strategy widely adopted in recent work \citep{lei2024planning, jiang2023self, islam-etal-2024-mapcoder}.

Repeated sampling and refinement iteratively produce new solutions based on the correctness of $s$ on training instances $I_r$, and validate $s$ on test instances $I_t$ once it passes $I_r$ or the iteration limit is reached. When $s := p$, its correctness is evaluated by querying the LLM to generate each output image $i^o$ given its corresponding input $i^i$ and the solution plan $p$, where $(i^i, i^o) \in I_r$ or  $(i^i, i^o) \in I_t$. Alternatively, when $s := c$, its correctness is assessed by executing $c$ on $I_r$ or $I_t$. In repeated sampling, the LLM iteratively generates a new plan $p$ and code $c$ from the problem description $Q$ without additional feedback. In contrast, refinement revises $p$ and $c$ by presenting the LLM with the previously incorrect $p$ and $c$, concatenated with failed training instances. In total, nine ARC solvers are employed to evaluate the performance of reasoning-oriented LLMs on the ARC benchmark, and the most competitive solver is treated as the ARC solver backbone.

\begin{figure*}[t]
 
\centering
\includegraphics[width=1\textwidth]{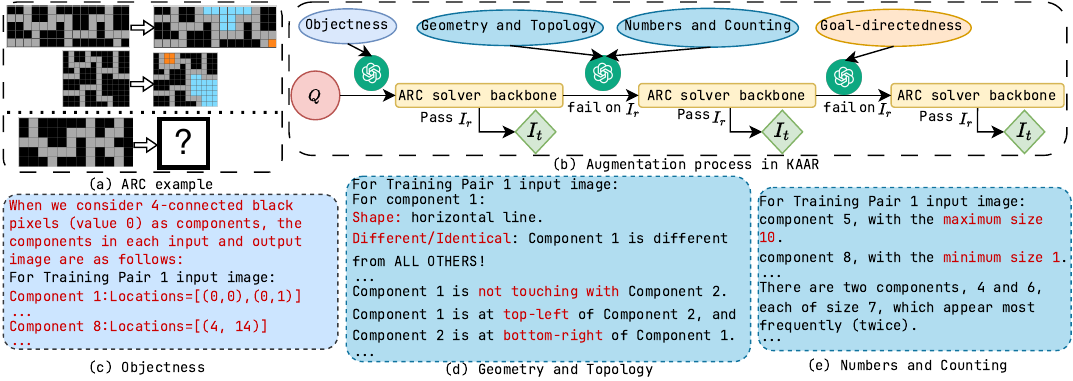}
\vspace{-0.6cm}
\caption{Augmentation process in KAAR (block (b)) and the corresponding knowledge augmentation fragments (blocks (c-e)) for ARC problem \textit{62ab2642} (block (a)).}
\vspace{-0.3cm}

\label{fig4}
\end{figure*}

\section{Knowledge Augmentation}

\textbf{Prior Knowledge Encoding}. 
The core knowledge priors specified in ARC can be formulated as a finite set of lifted, structured primitives (predicates and operators), enabling ARGA and GPAR to manually encode and instantiate them to ensure prior satisfaction. GPAR models predicates as abstraction-defined nodes enriched with attributes and inter-node relations, which are extracted using standard image processing algorithms.
Building on this insight, we propose KAAR, a knowledge augmentation approach for solving ARC tasks using reasoning-oriented LLMs. 
Following the core knowledge priors formulated in GPAR, KAAR expands their scope, revises their implementations, and categorizes these priors into four knowledge dimensions in ARC. In detail, KAAR adopts fundamental abstraction methods from GPAR to enable objectness. Objects are typically defined as components based on adjacency rules and color consistency (e.g., 4-connected or 8-connected components), while also including the entire image as a component. %KAAR further introduces additional abstractions: 
{KAAR further introduces four novel abstractions}: (1) \textit{middle-vertical}, which vertically splits the image into two equal parts, and treats each as a distinct component; (2) \textit{middle-horizontal}, which applies the same principle along the horizontal axis; (3) \textit{multi-lines}, which segments the image using full-length rows or columns of uniform color, and treats each resulting part as a distinct component; and (4) \textit{no abstraction}, which considers only raw 2D matrices. Under \textit{no abstraction},  KAAR degrades to the ARC solver backbone without incorporating any priors. 
KAAR incorporates GPAR’s component attributes (size, color, shape) and relations (spatial, congruent, inclusive), treating them as geometric and topological priors, respectively. %It further extends the attribute set with symmetry, bounding box, nearest boundary, and hole count, and augments the relation set with touching.
{It further extends with four novel attributes, including symmetry, bounding box, nearest boundary, and hole count, as well as a new touching relation.} For numeric and counting priors, KAAR adopts GPAR’s largest/smallest component sizes, and the most/least frequent component colors, while extending them with novel statistical analysis of hole counts,  symmetry and color diversity, as well as the most/least frequent sizes and shapes.

\setlength{\intextsep}{0pt} %
\begin{figure}
    \centering
    \includegraphics[width=0.4\textwidth]{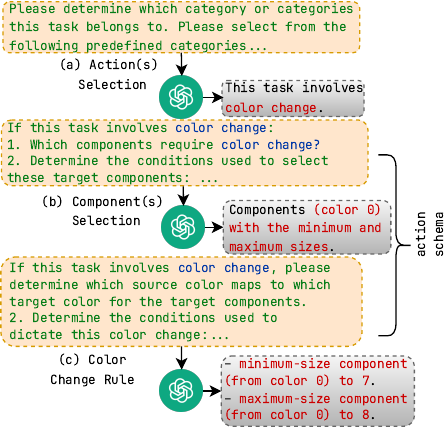} 
    \caption{Goal-directedness priors augmentation with input and response fragments from GPT-o3-mini. }
    \label{fig3}
\end{figure}

GPAR approaches goal-directedness priors by searching for a sequence of program instructions \citep{lei2023novelty} defined in a DSL. Each instruction supports conditionals, branching, looping, and action statements. %KAAR incorporates the condition and action concepts from GPAR, 
Inspired by these condition and action concepts, KAAR novelly enables goal-directedness priors by augmenting LLM knowledge in two steps: 1) It prompts the LLM to identify the most relevant actions for solving the given ARC problem from ten predefined action categories (Figure \ref{fig3} block (a)), partially derived from GPAR and extended based on the training set, such as color change, movement, and extension; 2) For each selected action, KAAR instructs the LLM with the associated schema to resolve implementation details. For example, for a color change action, KAAR first queries the LLM to identify the target components (Figure \ref{fig3} blocks (b)), and then specify the source and target colors for modification based on the target components (Figure \ref{fig3} blocks (c)). We note that KAAR also instructs the LLM to incorporate condition-aware reasoning when determining action implementation details, using knowledge derived from geometry, topology, numbers, and counting priors. This enables fine-grained control, for example, applying color changes only to black components conditioned on the maximum or minimum size: from black (value 0) to blue (value 8) if largest, or to orange (value 7) if smallest.  Figure~\ref{fig3} shows fragments of the goal-directedness prior augmentation. See Appendix~\ref{kaar_priors} for whole priors in KAAR.

\textbf{Ontology Construction}. In contrast to GPAR and previous work that leave priors unorganized \cite{xu2022graphs,tan2024llms},
KAAR structures the full set of core knowledge priors
into an ontology, where priors are organized into three hierarchical levels based on their dependencies. KAAR augments LLMs with priors at each level to enable incremental augmentation via in-context learning. This reduces context interference and supports stage-wise reasoning aligned with human cognitive development \citep{babakr2019piaget}. Figure~\ref{fig4}, block (b), illustrates the augmentation process in KAAR alongside the augmented prior fragments used to solve the problem shown in block (a). KAAR begins augmentation with objectness priors, encoding images into components with detailed coordinates based on a specific abstraction method  (block (c)). KAAR then prompts geometry and topology priors (block (d)), followed by numbers and counting priors (block (e)). These priors reside at the same ontological level, as they all build upon objectness. Finally, KAAR augments goal-directedness priors, as shown in Figure~\ref{fig3}, where target components are derived from objectness analysis and conditions are inferred from geometric, topological, and numerical analyses. After augmenting each level of priors, KAAR invokes the ARC solver backbone to generate solutions. If any solution passes $I_r$, it is validated on $I_t$; otherwise, augmentation proceeds to the next level of priors.

\textbf{Prior Restriction}. While the ontology provides a hierarchical representation of priors, it may also introduce hallucinations, such as duplicate abstractions, irrelevant component attributes or relations, and inapplicable actions. To address this, KAAR integrates additional restrictions to filter out inapplicable priors. KAAR adopts the duplicate-checking strategy in GPAR, retaining only abstractions that yield distinct components by size, color, or shape, in at least one training instance. In KAAR, each abstraction is associated with a  set of applicable priors.  For instance, when the entire image is treated as a component, relation priors are excluded, and actions such as movement and color change are omitted, whereas symmetry and size attributes are retained and actions such as flipping and rotation are considered.  In contrast, 4-connected and 8-connected abstractions include all component attributes and relations, and the full set of ten action priors. See Appendix~\ref{kaar_restriction} for detailed restrictions.

\section{Experiments} \label{sec:experiments}

\textbf{Experimental Setup}. We begin by comparing nine candidate solvers on the full ARC public evaluation set of 400 tasks. This offers broader insights than previous studies limited to the subsets of 400 training tasks \citep{lei2024generalized, xu2022graphs, wang2024hypothesis}, given the greater difficulty of the evaluation set \citep{legris2024h}. We experiment with  reasoning-oriented LLMs, including proprietary models, 
GPT-o3-mini \citep{o3miniintroducing}, and Gemini-2.0-Flash-Thinking (Gemini-2.0) \citep{Gemini}, and open-source models, DeepSeek-R1-Distill-Llama-70B (DeepSeek-R1-70B) \citep{guo2025deepseek} and QwQ-32B \citep{QwQ_32B}. We compute accuracy on test instances $I_t$ as the primary evaluation metric. It measures the proportion of problems where the first solution passes training instances $I_r$ and successfully solves $I_t$; otherwise, if none pass $I_r$ within 12 iterations, the last solution is evaluated on $I_t$, applied to both repeated sampling and refinement strategies. We also report accuracy on $I_r$ and $I_r\&I_t$, measuring the percentage of problems whose solutions solve $I_r$ and both $I_r$ and $I_t$. See Appendix \ref{Parameter} for parameter settings. 

\definecolor{LightBlue}{RGB}{232,243,255}
\begin{table*}[t]
\small
\setlength{\tabcolsep}{8.3pt}
\renewcommand{\arraystretch}{1}
\begin{tabular}{lcccccccccc}
\hlineB{2}
                & &\multicolumn{3}{c}{Direct Generation}       & \multicolumn{3}{c}{Repeated Sampling}       & \multicolumn{3}{c}{Refinement}              \\
                                  \noalign{\vskip -2pt}
               \cmidrule(lr){3-5}\cmidrule(lr){6-8}\cmidrule(lr){9-11}
                                 \noalign{\vskip -2pt}
                && \textit{P} & \textit{C} & \textit{PC}    & \textit{P} & \textit{C} & \textit{PC}     & \textit{P} & \textit{C} & \textit{PC}   \\           \noalign{\vskip -2pt}\cmidrule(lr){3-3} \cmidrule(lr){4-5} \cmidrule(lr){6-6} \cmidrule(lr){7-8} \cmidrule(lr){9-9} \cmidrule(lr){10-11} 
                  \hline 
\multirow{3}{*}{GPT-o3-mini}    & $I_r$& -         & -       & -          & 35.50      & \textbf{52.50}       & 35.50          & 31.00         & 47.25      & 32.00          \\
& \cellcolor{LightBlue}$I_t$ & \cellcolor{LightBlue}20.50       & \cellcolor{LightBlue}24.50       & \cellcolor{LightBlue}22.25          & \cellcolor{LightBlue}23.75      & \cellcolor{LightBlue}\textbf{\textcolor{red}{32.50}}       & \cellcolor{LightBlue}30.75          & \cellcolor{LightBlue}24.75         & \cellcolor{LightBlue}29.25      & \cellcolor{LightBlue}25.75          \\ 
& $I_r\&I_t$ & -      & -      & -         & 22.00      & \textbf{{31.75}}       & 29.25          & 21.75         & 28.50      & 25.00          \\ \hline

\multirow{3}{*}{Gemini-2.0}    & $I_r$ & -          & -         & -           & 36.50            & \textbf{39.50}      & 21.50          & 15.50          & 25.50         & 15.50          \\
 & \cellcolor{LightBlue}$I_t$ & \cellcolor{LightBlue}7.00             & \cellcolor{LightBlue}6.75       & \cellcolor{LightBlue}6.25           & \cellcolor{LightBlue}10.00            & \cellcolor{LightBlue}14.75      & \cellcolor{LightBlue}\textbf{\textcolor{red}{16.75}}          & \cellcolor{LightBlue}8.75          & \cellcolor{LightBlue}12.00         & \cellcolor{LightBlue}11.75          \\
 
  & $I_r\&I_t$ & -             & -       & -          & 9.50            & 14.25      & \textbf{{16.50}}          & 8.00          & 10.50         & 10.75          \\\hline
  
\multirow{2}{*}{QwQ-32B}       & $I_r$ & -           & -       & -           & \textbf{19.25}         & 13.50       & 15.25          & 16.75            & 15.00      & 14.25            \\
& \cellcolor{LightBlue}$I_t$ & \cellcolor{LightBlue}9.50           & \cellcolor{LightBlue}7.25       & \cellcolor{LightBlue}5.75           & \cellcolor{LightBlue}11.25         & \cellcolor{LightBlue}13.50       & \cellcolor{LightBlue}\textbf{\textcolor{red}{14.25}}          & \cellcolor{LightBlue}11.00            & \cellcolor{LightBlue}\textbf{\textcolor{red}{14.25}}     & \cellcolor{LightBlue}14.00            \\

& $I_r\&I_t$ & -          & -       & -           &  9.25       & 12.75     & \textbf{13.00}          & 8.75            & \textbf{{13.00}}     & 11.75            \\\hline

\multirow{2}{*}{DeepSeek-R1-70B} &$I_r$ & -          & -       & -            & \textbf{8.75}          & 6.75       & 7.75           & 6.25           & 5.75       & 7.75 \\
&\cellcolor{LightBlue}$I_t$ & \cellcolor{LightBlue}4.25          & \cellcolor{LightBlue}4.75       & \cellcolor{LightBlue}4.50            & \cellcolor{LightBlue}4.25          & \cellcolor{LightBlue}7.25       & \cellcolor{LightBlue}\textbf{\textcolor{red}{7.75}}           & \cellcolor{LightBlue}4.75           & \cellcolor{LightBlue}5.75       & \cellcolor{LightBlue}\textbf{\textcolor{red}{7.75}} \\
& $I_r\&I_t$ & -         & -      & -           & 3.50          & 6.50       &\textbf{7.25}           & 4.25           & 5.25       & 7.00 \\
\hlineB{2}
\end{tabular} 
\caption{Performance of nine ARC solvers measured by accuracy on $I_r$, $I_t$, and $I_r\&I_t$ using four reasoning-oriented LLMs. For each LLM, the highest accuracy on $I_r$ and $I_r\&I_t$ is in bold; the highest accuracy on $I_t$ is  in red. Accuracy is reported as a percentage. \textit{P} denotes the solution plan; \textit{C} and \textit{PC} refer to standalone and planning-aided code generation, respectively.}
\label{table1}
\end{table*}

\textbf{Results on Program Synthesis}. Table~\ref{table1} reports the performance of nine ARC solvers across four reasoning-oriented LLMs. For direct generation methods, accuracy on $I_r$ and $I_r\&I_t$ is omitted, as solutions are evaluated directly on $I_t$. GPT-o3-mini outperforms all other LLMs, achieving the highest accuracy on $I_r$ (52.50\%), $I_t$ (32.50\%), and $I_r\&I_t$ (31.75\%) under repeated sampling with standalone code generation (\textit{C}), highlighting its strong abstract reasoning capabilities. Notably, QwQ-32B, the smallest model, outperforms DeepSeek-R1-70B across all solvers and surpasses Gemini-2.0 under refinement. Among nine solvers, repeated sampling-based methods generally outperform those based on direct generation or refinement. This diverges from previous findings where refinement dominated conventional code generation tasks that lack abstract reasoning and generalization demands \citep{lei2024planning, zhong2024ldb, chen2023teaching} (see Appendix \ref{Performance Analysis} for detailed analysis). Within repeated sampling, planning-aided code generation (\textit{PC}) yields the highest accuracy on $I_t$ across most LLMs. It also demonstrates the strongest generalization with GPT-o3-mini and Gemini-2.0, as evidenced by the smallest accuracy gap between $I_r$ and $I_r\&I_t$.
A similar trend is observed on QwQ-32B and DeepSeek-R1-70B, where both \textit{C} and \textit{PC} generalize effectively across repeated sampling and refinement. Overall, repeated sampling with planning-aided code generation (RSPC) shows the best performance and thus serves as the ARC solver backbone. 

\definecolor{LightBlue}{RGB}{232,243,255}
\begin{table}[t]
\resizebox{\columnwidth}{!}{
\small
\setlength{\tabcolsep}{2pt}
\renewcommand{\arraystretch}{1.2}
\begin{tabular}{llccccccccc}
\hlineB{2}
                &      & \multicolumn{3}{c}{$I_r$}   & \multicolumn{3}{c}{$I_t$} &\multicolumn{3}{c}{$I_r\&I_t$}  \\
              \noalign{\vskip -2pt}
               \cmidrule(lr){3-5}\cmidrule(lr){6-8}\cmidrule(lr){9-11}
                                 \noalign{\vskip -2pt}
                &      & Acc   & $\Delta$ & $\gamma$ & Acc   & $\Delta$ & $\gamma$ & Acc   & $\Delta$ & $\gamma$ \\ \hline\hline
\multirow{2}{*}{\makecell{GPT-o3-mini}}     & RSPC   & 35.50 & -        & -        & 30.75 & -        & -     & 29.25 & -        & -     \\
                &\cellcolor{LightBlue}KAAR & \cellcolor{LightBlue}\textbf{\textcolor{red}{40.00}}   &\cellcolor{LightBlue} 4.50     & \cellcolor{LightBlue}12.68    & \cellcolor{LightBlue}\textbf{\textcolor{red}{35.00}} &\cellcolor{LightBlue} 4.25      & \cellcolor{LightBlue}13.82    & \cellcolor{LightBlue}\textbf{{\textcolor{red}{33.00}}} &\cellcolor{LightBlue} 3.75      & \cellcolor{LightBlue}12.82   \\ \hline
\multirow{2}{*}{\makecell{Gemini-2.0}}      & RSPC   & 21.50  & -        & -        & 16.75 & -        & -   & 16.50 & -        & -       \\
                & \cellcolor{LightBlue}KAAR &\cellcolor{LightBlue} 25.75 & \cellcolor{LightBlue}4.25     & \cellcolor{LightBlue}19.77    & \cellcolor{LightBlue}21.75 & \cellcolor{LightBlue}5.00        &\cellcolor{LightBlue} 29.85   & \cellcolor{LightBlue}20.50 & \cellcolor{LightBlue}4.00        &\cellcolor{LightBlue} 24.24   \\ \hline
\multirow{2}{*}{\makecell{QwQ-32B}}         & RSPC   & 15.25 & -        & -        & 14.25 & -        & - & 13.00 & -        & -       \\
                & \cellcolor{LightBlue}KAAR &\cellcolor{LightBlue} 22.25 & \cellcolor{LightBlue}\textbf{{7.00}}       & \cellcolor{LightBlue}45.90    & \cellcolor{LightBlue}21.00 & \cellcolor{LightBlue}\textbf{6.75}     & \cellcolor{LightBlue}47.37  & \cellcolor{LightBlue}19.25 & \cellcolor{LightBlue}\textbf{6.25}     & \cellcolor{LightBlue}48.08  
                \\ \hline
\multirow{2}{*}{\makecell{DeepSeek-R1-70B}} & RSPC   & 7.75  & -        & -        & 7.75  & -        & - & 7.25  & -        & -        \\
                & \cellcolor{LightBlue}KAAR & \cellcolor{LightBlue}12.25 & \cellcolor{LightBlue}4.50      & \cellcolor{LightBlue}\textbf{{58.06}}    & \cellcolor{LightBlue}12.75    & \cellcolor{LightBlue}5.00     & \cellcolor{LightBlue}\textbf{64.52} & \cellcolor{LightBlue}11.50    & \cellcolor{LightBlue}4.25     & \cellcolor{LightBlue}\textbf{{58.62}}
                \\
                \hlineB{2}
\end{tabular}
}
\caption{Comparison of RSPC and KAAR in terms of accuracy (Acc) on $I_r$, $I_t$, and $I_r\&I_t$. $\Delta$ and $\gamma$ denote the absolute and relative improvements over RSPC, respectively. All values are reported as percentages. For $I_r$, $I_t$, and $I_r\&I_t$, the highest Acc values are in red and the  best $\Delta$ and $\gamma$ results are in bold.}

\label{table2}
\end{table}

\textbf{Results on KAAR}. We further compare the performance of RSPC with its knowledge-augmented variant, KAAR. For each task, KAAR begins with simpler abstractions, i.e., no abstraction and whole image, and progresses to complicated 4-connected and 8-connected abstractions, consistent with GPAR. KAAR reports the accuracy on test instances $I_t$ based on the first abstraction whose solution solves all training instances $I_r$; otherwise, it records the final solution from each abstraction and selects the one that passes the most $I_r$ to evaluate on $I_t$ (details in Appendix~\ref{KAAR_Details}).
KAAR allows the solver backbone (RSPC) up to 4 iterations per invocation, totaling 12 iterations, consistent with the non-augmented setting. As shown in Table~\ref{table2}, KAAR consistently outperforms non-augmented RSPC, yielding around 5\% absolute accuracy gains on $I_r$, $I_t$, and $I_r\&I_t$ across all LLMs. This highlights the effectiveness and model-agnostic nature of the augmented priors. KAAR achieves the highest accuracy using GPT-o3-mini, i.e., 40\% on $I_r$, 35\% on $I_t$, and 33\% on $I_r\&I_t$. It shows the best absolute improvements ($\Delta$), around 6.5\%, using QwQ-32B and the most pronounced  relative gains ($\gamma$), over 58\%, using DeepSeek-R1-70B on $I_r$, $I_t$, and $I_r\&I_t$. 
Moreover, KAAR maintains generalization comparable to RSPC across  all LLMs, indicating that the augmented priors are sufficiently abstract and expressive to serve as basis functions for reasoning, in line with ARC assumptions. See Appendix~\ref{Generalization} for a discussion of generalization and Appendix~\ref{Additional_Experimental_Results} for evaluations on additional ARC-related and non-ARC benchmarks.

\setlength{\intextsep}{0pt} %
\begin{figure}
    \centering
    \includegraphics[width=0.47\textwidth]{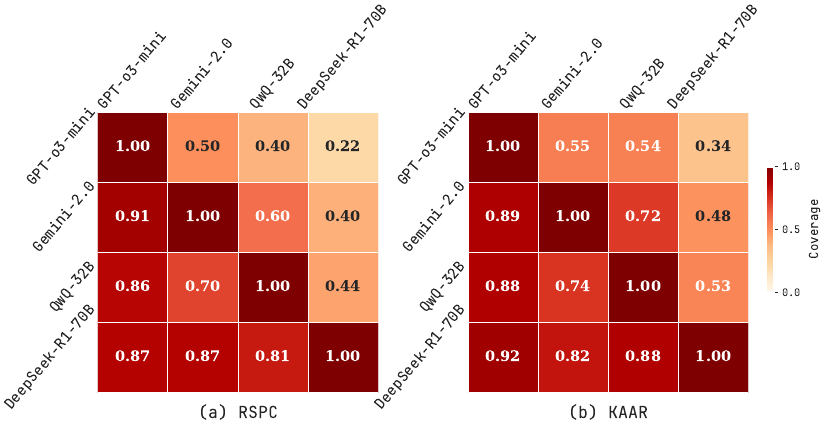}
    \caption{Asymmetric relative coverage matrices for RSPC (a) and KAAR (b), showing the proportion of problems whose test instances are solved by the row model that are also solved by the column model.}
    
    \label{heatmap}
\end{figure}

\textbf{Performance Analysis}. We compare relative problem coverage across evaluated LLMs under RSPC and KAAR based on successful solutions on test instances. As shown in Figure~\ref{heatmap}, each cell $(i,j)$ is computed as $\frac{|A_i \cap A_j|}{|A_i|}$, where $A_i$ and $A_j$ are the sets of problems solved by the row and column LLMs, respectively. Values near 1 indicate that the column LLM covers most problems solved by the row LLM. Under RSPC (Figure~\ref{heatmap} (a)), GPT-o3-mini exhibits broad coverage, with column values consistently above 0.85. Gemini-2.0 and QwQ-32B also show substantial alignment, with mutual coverage exceeding 0.6. Figure~\ref{heatmap} (b) illustrates that KAAR generally improves or maintains inter-model overlap compared to RSPC. Notably, the coverage between GPT-o3-mini and DeepSeek-R1-70B increases from 0.22 under RSPC to 0.34 under KAAR. These results underscore the effectiveness of KAAR in improving cross-model generalization, enabling all evaluated LLMs to solve additional shared problems and allowing smaller models such as QwQ-32B and DeepSeek-R1-70B to better align with stronger LLMs on ARC. See Appendix \ref{solution_coverage} for problem coverage across evaluated solvers.

\setlength{\intextsep}{0pt} %
\begin{figure}
    \centering
    \includegraphics[width=0.47\textwidth]{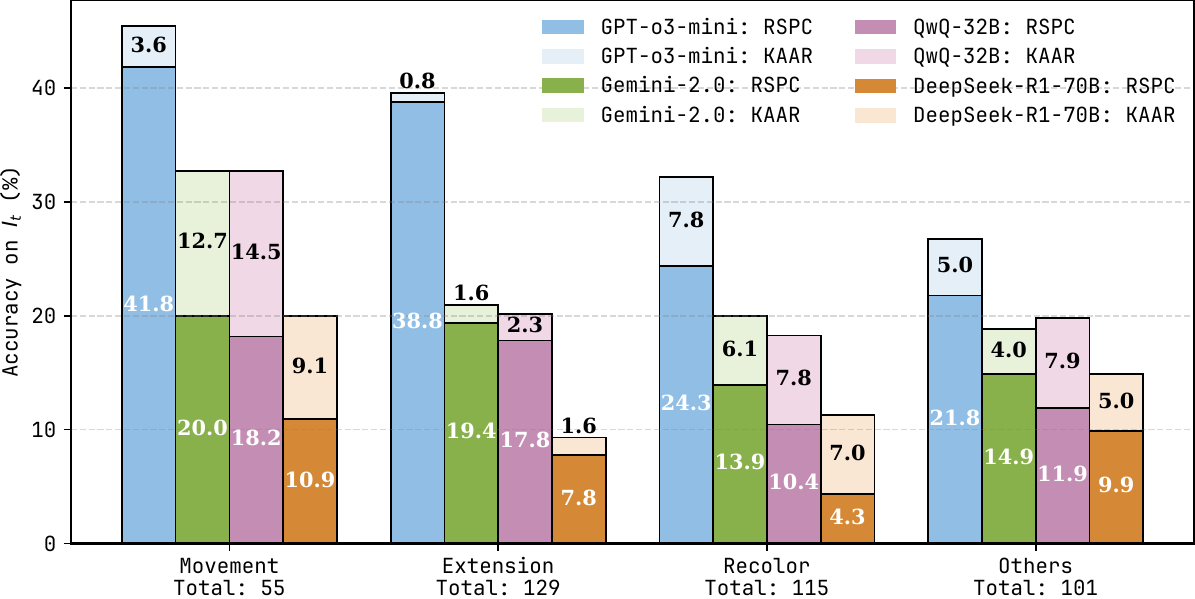}
    \caption{Accuracy on test instances $I_t$ for RSPC and KAAR across the \textit{movement}, \textit{extension}, \textit{recolor}, and \textit{others} categories using four LLMs.  Each stacked bar shows RSPC accuracy (darker segment) and the additional improvement from KAAR (lighter segment).}
    \label{class_percentage}
\end{figure}

\textbf{Performance by Category}. Following prior work \cite{xu2022graphs,lei2024generalized}, we categorize the 400 problems in the ARC public evaluation set into four classes based on their primary transformations: (1) \textit{movement} (55 problems), (2) \textit{extension} (129 problems), (3) \textit{recolor} (115 problems), and (4) \textit{others} (101 problems). 
The \textit{others} category comprises infrequent transformations. See Appendix~\ref{Categories} for examples of each category. 
Figure~\ref{class_percentage} illustrates the accuracy on test instances $I_t$ for RSPC and KAAR across four categories with evaluated LLMs. Each stacked bar represents RSPC accuracy and the additional improvement contributed by KAAR. KAAR consistently outperforms RSPC with the largest accuracy gain in \textit{movement} (14.5\% with QwQ-32B). 
It shows limited improvements in \textit{extension}, where several problems involve pixel-level extension, which minimizes the reliance on component-level recognition. 
Moreover, \textit{extension} requires accurate spatial inference across multiple components. This poses greater difficulty than \textit{movement}, which requires mainly direction identification. {We note that multi-component spatial inference remains a persistent challenge for LLMs \citep{yamada2024evaluating,cohn2023dialectical,bang-etal-2023-multitask}, highlighting a difficulty that is intrinsic to ARC and reinforcing its role as a challenging benchmark for abstract reasoning and generalization.}

\setlength{\intextsep}{0pt} %
\begin{figure}
    \centering
    \includegraphics[width=0.47\textwidth]{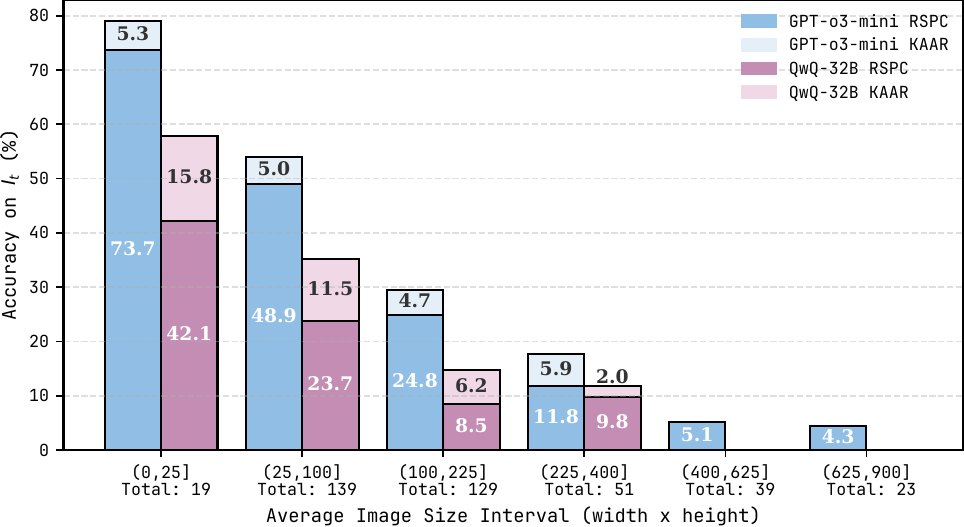}
    \caption{Accuracy on test instances $I_t$ for RSPC and KAAR across average image size intervals, evaluated using GPT-o3-mini and QwQ-32B. See Figure~\ref{fig:comparsion_other_two} in Appendix for results on other LLMs.}
        
    \label{fig5}
\end{figure}

\textbf{Impact of Size}. A notable feature of ARC is the variation in image size both within and across problems. We categorize tasks by averaging the image size per problem, computed over both training and test image pairs. We report the accuracy on $I_t$ for RSPC and KAAR across average image size intervals using GPT-o3-mini and QwQ-32B, the strongest proprietary and open-source models. As shown in Figure~\ref{fig5}, both LLMs experience performance degradation as image size increases. When the average image size exceeds 400 pixels (20×20), GPT-o3-mini solves only three problems, while QwQ-32B solves none. Benefiting from object-centric representations, KAAR consistently outperforms RSPC on problems with average image sizes below 400 pixels. By abstracting each image into components, KAAR reduces interference from irrelevant pixels, directs attention to salient components, and facilitates component-level transformation analysis. However, larger images tend to yield oversized and numerous components after abstraction, where oversized components impede transformation execution and numerous components complicate the identification of target components.

\setlength{\intextsep}{0pt} %
\begin{figure}
    \centering
    \includegraphics[width=0.47\textwidth]{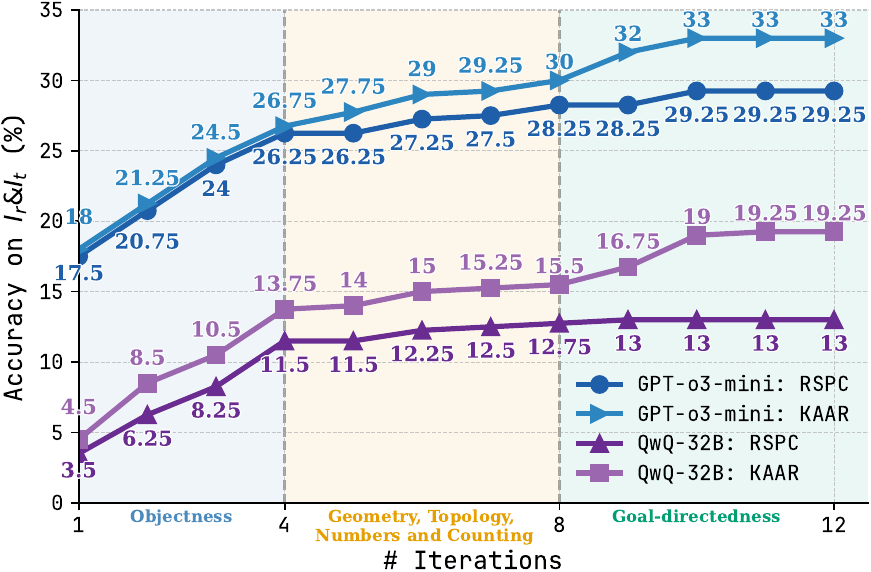}

    \caption{Accuracy variance on $I_r\&I_t$ over iterations for RSPC and KAAR using GPT-o3-mini and QwQ-32B. See  Appendix Figure~\ref{fig:performance_with_known_variance_the_other_two} for results on other LLMs.}
        
    \label{fig6}
\end{figure}

\textbf{Scaling with Iterations}. Figure~\ref{fig6} presents the variance in accuracy on $I_r\&I_t$ for RSPC and KAAR as iteration count increases using GPT-o3-mini and QwQ-32B. For each task under KAAR, we record iterations from the abstraction that solves both $I_r$ and $I_t$. For KAAR, performance improvements across each 4-iteration block are driven by the solver backbone invocation after augmenting an additional level of priors. RSPC shows rapid improvement in the first 4 iterations and plateaus around iteration 8. KAAR consistently outperforms RSPC, with the performance gap progressively increasing after additional priors are augmented and peaking after the integration of goal-directedness.

\begin{table}[t]
\centering
\small
\setlength{\tabcolsep}{5pt}
\renewcommand{\arraystretch}{1.2}
\resizebox{\columnwidth}{!}{
\begin{tabular}{llc@{\hspace{8pt}}cc@{\hspace{8pt}}cc}
\hlineB{2}
& & KAAR
& \multicolumn{2}{c}{KAAR-P}
& \multicolumn{2}{c}{KAAR-S} \\
\cmidrule(lr){3-3}\cmidrule(lr){4-5}\cmidrule(lr){6-7}
& & Acc & Acc & $\Delta$ & Acc & $\Delta$ \\
\hline\hline

\multirow{3}{*}{Gemini-2.0}
& $I_r$
& 25.75
& 22.25 & -3.50
& 23.00 & -2.75 \\
& \cellcolor{LightBlue}$I_t$
& \cellcolor{LightBlue}21.75
& \cellcolor{LightBlue}18.75
& \cellcolor{LightBlue}-3.00
& \cellcolor{LightBlue}19.00
& \cellcolor{LightBlue}-2.75 \\
& $I_r\&I_t$
& 20.50
& 17.75 & -2.75
& 18.00 & -2.50 \\
\hline

\multirow{3}{*}{QwQ-32B}
& $I_r$
& 22.25
& 18.75 & -3.50
& 18.50 & -3.75 \\
& \cellcolor{LightBlue}$I_t$
& \cellcolor{LightBlue}21.00
& \cellcolor{LightBlue}17.25
& \cellcolor{LightBlue}-3.75
& \cellcolor{LightBlue}17.75
& \cellcolor{LightBlue}-3.25 \\
& $I_r\&I_t$
& 19.25
& 16.00 & -3.25
& 16.25 & -3.00 \\
\hline

\multirow{3}{*}{DeepSeek-R1-70B}
& $I_r$
& 12.25
& 11.25 & -1.00
& 9.00 & -3.25 \\
& \cellcolor{LightBlue}$I_t$
& \cellcolor{LightBlue}12.75
& \cellcolor{LightBlue}11.00
& \cellcolor{LightBlue}-1.75
& \cellcolor{LightBlue}9.00
& \cellcolor{LightBlue}-3.75 \\
& $I_r\&I_t$
& 11.50
& 11.00 & -0.50
& 8.50 & -3.00 \\
\hlineB{2}

\end{tabular}
}
\caption{Accuracy (Acc) on $I_r$, $I_t$, and $I_r\&I_t$ for KAAR and its variants across three LLMs. KAAR-P removes the newly introduced priors and retains only the priors from GPAR. KAAR-S invokes the solver backbone (RSPC) only after all knowledge priors are augmented. $\Delta$ denotes the accuracy difference relative to KAAR. All values are reported as percentages.}

\label{tab:kaar_ablation}
\end{table}

\textbf{Ablation Study}. KAAR adopts the core-knowledge priors in GPAR and extends them with additional priors required for solving ARC tasks. These priors are organized within a dependency-aware ontology that facilitates progressive knowledge augmentation and stage-wise reasoning.  We evaluate two variants, KAAR-P and KAAR-S. KAAR-P excludes the newly introduced priors while preserving the priors from GPAR and progressive knowledge augmentation with stage-wise reasoning. KAAR-S removes progressive augmentation and stage-wise reasoning. Unlike KAAR, which invokes the solver backbone (RSPC) after augmenting each level of priors, KAAR-S uses RSPC to solve each problem within 12 iterations after augmenting all priors at once.

Table~\ref{tab:kaar_ablation} reports the results for the same reasoning-oriented LLMs as in Tables~\ref{table1} and~\ref{table2}, excluding GPT-o3-mini due to its computational cost. KAAR-P and KAAR-S show decreased accuracy across all evaluation metrics, $I_r$, $I_t$, and $I_r\&I_t$, for all evaluated LLMs, demonstrating the effectiveness of the newly introduced priors in KAAR and progressive knowledge augmentation with stage-wise reasoning. Presenting all knowledge priors simultaneously introduces superfluous information, which may obscure viable solutions and impair LLM reasoning accuracy.  Appendix~\ref{ablation} further analyzes the influence of the dependency-aware augmentation order defined by the ontology.

\section{Discussion}

\textbf{Rationale for ARC}. While related benchmarks exist, ARC remains a highly authentic measure of generalization. In contrast to RAVEN \citep{zhang2019raven} and PGM \citep{barrett2018measuring}, which primarily evaluate closed-form rule fitting within predefined rule spaces, ARC assumes that tasks are solvable through core knowledge priors, while intentionally leaving the problems underspecified to prevent the explicit encoding of complete solution rules \citep{chollet2019measure}.  
Each ARC task is unique, defining a distinct domain that requires solvers to infer solutions from a few examples and generalize to test cases, thereby advancing models beyond rule fitting toward genuine generalization. Recent training approaches that employ data augmentation to generate additional examples for each task \citep{akyurek2024surprising,franzen2025product} achieve notable results on ARC. However, they deviate substantially from ARC’s objective of solving each task from limited examples without reliance on externally generated data \cite{chollet2019measure}.
 
\textbf{Cost Analysis}. {As the solver backbone, RSPC uses fewer tokens than its refinement-based counterpart while achieving comparable or higher accuracy.} In KAAR, knowledge augmentation increases token consumption, while the additional tokens remain relatively constant since all priors, except goal-directedness, are generated via image processing algorithms. On GPT-o3-mini, augmentation tokens constitute around 60\% of solver backbone token usage, while on QwQ-32B, this overhead decreases to about 20\%, as the solver backbone consumes more tokens.  See Appendix~\ref{token_analysis} for a detailed discussion.

\textbf{Solution Analysis}.  
To assess adherence to core knowledge priors, we manually reviewed the solution plans and code generated by RSPC and KAAR that successfully solve $I_t$ with GPT-o3-mini. RSPC tends to solve problems without object-centric reasoning. For instance, in Figure~\ref{fig1}, it shifts each row downward by one and pads the top with zeros, rather than reasoning over objectness to move each 4-connected component down by one step. We note that object recognition in ARC involves grouping pixels into task-specific components based on clustering rules, differing from feature extraction approaches \citep{zhao2019object} in conventional computer vision tasks. It remains challenging for current LLMs \citep{xu2024llms,li2024tackling}. In contrast, KAAR enables objectness through explicitly defined abstractions, implemented via standard image processing algorithms, thus ensuring both accuracy and generalization. KAAR-generated solutions are consistent with the augmented priors, as shown in Figure~\ref{fig3}.

\section{Conclusion}

We explored the generalization and abstract reasoning capabilities of reasoning-oriented LLMs on the ARC benchmark using nine candidate solvers. Experimental results show that repeated-sampling planning-aided code generation (RSPC) achieves the highest test accuracy and exhibits consistent generalization across most evaluated LLMs. To further improve performance, we propose KAAR, which progressively augments LLMs with core knowledge priors structured into hierarchical levels based on their dependencies, and applies RSPC after augmenting each level of priors to enable stage-wise reasoning. 
KAAR improves LLM performance on ARC while maintaining strong generalization compared to RSPC. However, ARC remains challenging even for the most capable reasoning-oriented LLMs, given its emphasis on  abstract reasoning and generalization, highlighting current limitations and motivating future research.

\section*{Limitations} \label{limitations}

KAAR improves the performance of reasoning-oriented LLMs on ARC tasks by progressively prompting with core knowledge priors. Although this inevitably increases token usage, the trade-off can be justified, as the exploration of LLM generalization remains in its early stages. KAAR integrates diverse abstraction methods to enable objectness and iteratively applies abstractions in order of increasing complexity. In contrast, humans typically infer appropriate abstractions directly from training instances, rather than leveraging exhaustive search. To address this, we prompt different LLMs with raw 2D matrices of each ARC problem to select one or three relevant abstractions, but the results are unsatisfactory. {Accurate abstraction inference often depends on validation through viable solutions, thereby shifting the challenge back to solution generation. Given the low solution-generation accuracy reported in Table \ref{table1}, direct abstraction inference remains a challenging task for evaluated LLMs.} Additionally, KAAR augments core knowledge priors through in-context learning but lacks mechanisms to enforce LLM adherence to these priors during reasoning. {However, the close correspondence between KAAR-generated solutions and augmented core knowledge priors (Figure \ref{fig3}) suggests a potential underlying alignment.}
Future work could explore fine-tuning or reinforcement learning to better align model behavior with the desired reasoning patterns.

\bibliography{custom_correct}
\clearpage

\appendix

\section*{Appendix}

\label{sec:appendix}

\section{Related Work} \label{related_work}

\textbf{Knowledge-Augmented LLMs.}  Augmenting LLMs with external knowledge can improve reasoning capabilities and mitigate hallucination in different tasks \citep{mialon2023augmented}. Previous studies achieve this by incorporating domain-specific knowledge, designed by human experts \citep{zhu-etal-2025-knowagent}, retrieved via search engines \citep{vu-etal-2024-freshllms}, or extracted from Wikipedia documents \citep{li2024chainofknowledge}. \citet{trivedi-etal-2023-interleaving} demonstrated that interleaving knowledge augmentation within reasoning steps further reduces model hallucination, resulting in more accurate multi-step reasoning. Additionally, augmenting LLMs with execution feedback improves performance on both question answering \citep{qiao-etal-2024-making} and program synthesis tasks \citep{lei2024generalized, zhong2024ldb, chen2023teaching}.  KAAR follows the interleaved knowledge augmentation approach and further introduces an ordered knowledge augmentation strategy based on knowledge hierarchical dependencies to mitigate context interference and facilitate stage-wise reasoning. Its superior performance over both the non-augmented variant (RSPC) and the non–stage-wise reasoning counterpart (KAAR-S) underscores its effectiveness.

\textbf{Search in DSL.} An abstract, expressive, and compositional representation of core knowledge priors is essential for solving ARC tasks \citep{chollet2019measure}. Recent studies have manually encoded core knowledge priors into domain-specific languages (DSLs) with lifted representations \citep{xu2022graphs, lei2024generalized, Kaggle2020}. Various program synthesis methods have been proposed to search for valid program solutions within their DSLs, including DAG-based search \citep{Kaggle2020}, graph-based constraint-guided search used in ARGA \citep{xu2022graphs}, and generalized planning  employed in GPAR \citep{lei2024generalized}.

KAAR benefits from the hand-crafted DSL in GPAR, where core knowledge priors are encoded with high precision and interpretability. Unlike GPAR, which performs an exhaustive search over the entire DSL, resulting in significant inefficiencies, KAAR leverages LLMs pretrained on extensive code-related corpora to synthesize programs. Moreover, the underlying code generation strategy in KAAR, RSPC, ensures high-quality and accurate program solutions. {KAAR further formalizes the core knowledge priors into a dependency-aware ontology, which preserves knowledge coverage and enables progressive knowledge augmentation for LLMs to support stage-wise reasoning, thereby reducing interference from irrelevant and noisy context.}

\textbf{LLMs for ARC.} Recent studies have explored using LLMs as ARC solvers to directly generate test output matrices and have provided LLMs with different problem descriptions to improve output accuracy. \citet{camposampiero2023abstract} employed LLMs to generate output grids from textual task descriptions, derived from a vision module which is designed to capture human-like visual priors. \citet{min2023approach} prompted LLMs with the raw 2D matrices of each task, along with transformation and abstraction examples. \citet{xu2024llms} demonstrated that object representations produced by predefined abstractions can improve LLM performance on ARC tasks. % Recent advances in code generation by LLMs \citep{lei2024planning, zhong2024ldb, chen2022codet} highlight their potential to replace search-based program synthesis,  addressing efficiency limitations.
Diverging from direct image generation, \citet{tan2024llms} evaluated LLM performance on the ARC benchmark by generating Python program solutions. Additionally, \citet{wang2024hypothesis} approached ARC as an inductive reasoning problem and introduced hypothesis search, where program solutions are generated by selecting LLM-generated hypotheses encoded as functions. 

When leveraging LLMs as ARC solvers, existing studies tend to emphasize accuracy on partial training set problems and overlook the core principle of ARC, where solutions should be constructed using core knowledge priors \citep{chollet2019measure}. However, LLMs still lack these priors, such as objectness, as evidenced by  RSPC-generated solutions.  %KAAR addresses this gap by progressively augmenting LLMs with structured core knowledge priors, alongside uniquely implemented goal-directedness priors. 
KAAR addresses this gap by progressively augmenting LLMs with structured core knowledge priors. KAAR achieves strong performance, 35.0\% test accuracy on the full evaluation set of 400 problems using GPT-o3-mini. It demonstrates substantial generalization, and produces solutions that adhere to core knowledge priors.

\textbf{Training-Based Methods.} 
The few training examples available per ARC task restrict the effectiveness of fine-tuning for achieving robust generalization and accurate reasoning in billion-parameter LLMs. To address this limitation, \citet{bikov2024reflection} fine-tuned LLMs on augmented ARC tasks constructed through standard image augmentation operations, including rotation, flipping, and permutation. %\citet{akyurek2024surprising} adopted test-time training to further improve LLM performance on ARC. In this paradigm, models are first fine-tuned on large-scale synthetic ARC datasets (RE-ARC)~\citep{hodel2024addressing}. For each ARC task, the LLM is subsequently trained on task-specific samples, including augmented data, original input–output pairs, and leave-one-out examples, to generate test outputs. 
\citet{akyurek2024surprising} applied test-time training to further improve LLM performance on ARC. Models are first fine-tuned on large-scale synthetic ARC data (RE-ARC) \citep{hodel2024addressing} and subsequently optimized for each ARC task using original input–output pairs, augmented examples, and leave-one-out examples. Extending this framework, \citet{franzen2025product} incorporated augmented data across test-time training, inference, and selection, following initial fine-tuning on RE-ARC, further advancing LLM performance on the ARC benchmark. \citet{hu2025arc} introduced a similar two-stage ViT training paradigm, with initial fine-tuning on RE-ARC followed by test-time training with canvas-derived augmented examples for each ARC task.

Training approaches have achieved state-of-the-art performance on ARC. However, they diverge from ARC’s original goal of evaluating human-like intelligence, which involves performing abstract reasoning and generalization from limited examples without reliance on externally generated data \citep{chollet2019measure}. Human solvers typically infer solutions solely from the provided examples \citep{johnson2021fast,acquaviva2021communicating}. Given this, training with additionally augmented data is inconsistent with ARC’s intent. Instead, learning core knowledge priors and performing inference grounded in them offers a  promising alternative. KAAR adopts this idea by augmenting LLMs through in-context learning with core knowledge priors, thereby preserving generalization while enhancing performance on ARC tasks.

\section{Core Knowledge Priors in KAAR} \label{kaar_priors}

KAAR incorporates abstractions to enable objectness priors; leverages component attributes, relations, and statistical analysis of component attributes to encode geometry, topology, numbers, and counting priors; and employs predefined actions to support goal-directedness priors. Table~\ref{abstraction} presents all abstractions used in KAAR, organized by their prioritization. KAAR incorporates fundamental abstractions, such as 4-connected and 8-connected components, from GPAR, and extends them with additional abstractions unique to KAAR, highlighted in red. Table~\ref{KAAR_Konwledge_prior} introduces geometry, topology, numbers, and counting priors, and ten predefined transformations used in KAAR. For each action, KAAR augments the LLM with its corresponding schema to resolve implementation details. The actions and their schemas are detailed in Table~\ref{KAAR_actions}. Most actions can be specified within three steps, keeping them tractable for LLMs.
\vspace{0.2cm}

\begin{figure*}[t]
    \centering
    \includegraphics[width=1\linewidth]{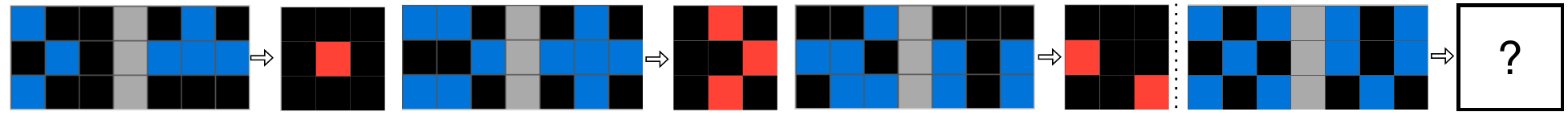}
    \caption{ARC problem \textit{0520fde7}}
    \label{ARC0520fde7}
\end{figure*}

\section{Restrictions in KAAR} \label{kaar_restriction}
For certain abstractions, some priors are either inapplicable or exclusive. The specific priors assigned to some abstractions are detailed in Table~\ref{restriection}. For the \textit{whole image} abstraction, few priors are applicable as the abstraction contains only a single component. In contrast, the \textit{4/8-connected-multi-color-non-background} abstractions retain most priors. The highlighted priors that capture per-component color diversity are used exclusively for \textit{4/8-connected-multi-color-non-background} abstractions, while priors tailored to a single-color component, such as \textit{components with same color}, \textit{components with most frequent color}, and \textit{components with least frequent color}, are excluded. For the \textit{middle-vertical} and \textit{middle-horizontal} abstractions, where the image is evenly divided into two components, flipping and movement actions are enabled to facilitate reasoning over  overlapping components. For instance, in the problem shown in Figure~\ref{ARC0520fde7}, the solution involves splitting the image along a middle-vertical grid line and moving one component to overlap the other. In the resulting component, a pixel is colored red if the overlapping pixels in both components are blue; otherwise, it is colored black.

\section{Parameter Settings}
\label{Parameter}
KAAR operates on all LLMs through API access with the full conversational history. For proprietary models, GPT-o3-mini and Gemini-2.0 Flash-Thinking (Gemini-2.0), we use the 2025-01-31 snapshot and the exp-1219 version, respectively, with default parameter settings. For open-source models, DeepSeek-R1-Distill-Llama-70B (DeepSeek-R1-70B) and QwQ-32B, we set temperature to 0.6, top-p to 0.95, and top-k to 40 to reduce repetitive outputs and filter rare tokens while preserving generation diversity.  We conduct experiments on a virtual machine with 4 NVIDIA A100  80GB GPUs. 

\section{Performance Analysis} \label{Performance Analysis}

In Table \ref{table1}, the incorporation of planning  facilitates accurate code generation under the repeated sampling setting.  For example, in Figure~\ref{15696249}, repeated sampling with planning-aided code generation produces a correct solution using GPT-o3-mini by replicating the input image horizontally or vertically based on the presence of a uniform row or column, as specified in the plan and implemented accordingly in code. In contrast, without planning assistance, standalone code generation produces incomplete logic, considering only whether the first column is uniform to determine the replication direction, which leads to failure on the test instance. 

For the ARC benchmark, repeated sampling–based methods achieve  higher accuracy on $I_r$, $I_t$, and $I_r\&I_t$ compared to refinement-based approaches when using GPT-o3-mini,  Gemini-2.0, and DeepSeek-R1-70B as shown in Table \ref{table1}. Figure~\ref{d19f7514} presents an ARC problem where repeated sampling with planning-aided code generation yields a correct solution, whereas its refinement variant fails to correct the initial erroneous code, and the flawed logic persists across subsequent refinements when using GPT-o3-mini.  Recent studies have shown that refinement struggles when the initial code significantly deviates from the intended functionality \citep{tian2023test}. Moreover, the lack of explicit correction instructions in feedback messages often leads to refinements diverging further from the intended solution \citep{lei2024planning}.

\section{KAAR} \label{KAAR_Details}

Algorithm~\ref{alg:kaar-details} presents the pseudocode of KAAR. For each abstraction, KAAR incrementally augments the LLM with core knowledge priors, structured into three dependency-aware levels: beginning with objectness (Line 5), followed by geometry and topology (Lines 10 and 12), numbers and counting (Line 14), and concluding with goal-directedness priors (Line 18). We note that KAAR encodes geometry and topology priors through component attributes (Line 9) and relations (Line 11). The full set of priors is detailed in Tables~\ref{abstraction},  ~\ref{KAAR_Konwledge_prior}, and ~\ref{KAAR_actions}. After augmenting each level of priors, KAAR invokes the solver backbone (RSPC) at Lines 6, 15, and 19 to generate code solutions guided by text-based plans, allowing up to 4 iterations (Lines 25–37). In each iteration, the solver backbone first validates the generated code on the training instances $I_r$; if successful, it then evaluates the solution on the test instances $I_t$.  The solver backbone returns \texttt{solve} if the generated solution successfully solves $I_t$ after passing $I_r$; \texttt{pass} if only $I_r$ is solved; or continues to the next iteration if the solution fails on $I_r$. If the solver backbone fails to solve $I_r$ within the allotted 4 iterations at Lines 6 and 15, KAAR augments the next level of priors. KAAR proceeds to the next abstraction when the solver backbone fails to solve $I_r$ at Line 19, after the 4-iteration limit.
KAAR terminates abstraction iteration upon receiving either \texttt{pass} or \texttt{solve} from the solver backbone and reports accuracy on $I_r$, $I_t$, and $I_r\&I_t$ accordingly. If no abstraction fully solves $I_r$, KAAR records the final code solution for each abstraction (Line 22), selects the one that passes the most training instances (Line 23), and evaluates it on $I_t$ to determine additional accuracy gains (Line 24).

KAAR generates priors offline using image processing algorithms at Lines 4, 9, 11 and 13. In contrast, KAAR enables goal-directedness priors at Line 18 by prompting the LLM to select the most suitable actions and identify their implementation details, as described in Table~\ref{KAAR_actions}. KAAR iterates over abstractions from simpler to more complex, following the order specified in Table~\ref{abstraction}. We note that the highest-priority abstraction is \textit{no abstraction}, where KAAR degrades to the solver backbone (RSPC) as no priors are applied.

\section{Generalization} \label{Generalization}

Table~\ref{table1} highlights performance variations across reasoning-oriented LLMs and ARC solvers with respect to both accuracy and generalization. Notably, the ARC solver, repeated sampling with standalone code generation, exhibits a substantial accuracy gap between $I_r$ and $I_r\&I_t$, indicating limited generalization capability when using GPT-o3-mini and Gemini-2.0. In contrast, repeated sampling with planning-aided code generation demonstrates markedly improved generalization by preventing solutions from directly replicating the output matrices of training instances, as illustrated in Figure~\ref{358ba94e}. This output copying, observed under repeated sampling with standalone code generation, accounts for approximately 24\% and 95\% of 83 and 101 overfitting problems with GPT-o3-mini and Gemini-2.0, respectively. When planning is incorporated, output copying is reduced to around 8\% and 35\% of 25 and 20 overfitting problems with GPT-o3-mini and Gemini-2.0, respectively. 

Planning-aided methods, RSPC and KAAR, demonstrate smaller accuracy gaps in Table~\ref{table2}. However, the  generalization challenge persists.  One reason is that solutions include low-level logic for the training pairs, thus failing to generalize. Another reason is the usage of incorrect abstractions. KAAR selects the first abstraction that solves $I_r$, to report accuracy on $I_t$, while this abstraction may overfit to $I_r$. Figures~\ref{ARC_695367ec} and~\ref{ARC_b1fc8b8e} illustrate two ARC problems, \textit{695367ec} and \textit{b1fc8b8e}, where both RSPC and KAAR successfully solve the training instances $I_r$ but fail on the test instances $I_t$ when using GPT-o3-mini. For problem \textit{695367ec}, the correct solution involves generating a fixed 15×15 output image by repeatedly copying the input image, changing its color to black, and adding internal horizontal and vertical lines colored with the original input image’s color. However, the RSPC-generated code applies a distinct rule to each input image size without considering generalization. For problem \textit{b1fc8b8e}, the solution requires accurate object recognition despite component contact, and correctly placing each component into one of the four corners. However, RSPC fails to recognize objectness, and its solution deviates from human intuition, being overfitted to $I_r$. For problems \textit{695367ec} and \textit{b1fc8b8e}, KAAR exhibits the same limitations, although it adopts abstractions to enable objectness. KAAR begins with the simplest abstraction, \textit{no abstraction}, where KAAR degrades to RSPC. As a result, it generates the same solution as RSPC and terminates without attempting other abstractions, since the solution already solves $I_r$ and is then evaluated on $I_t$, resulting in overfitting.

\begin{table*}[t]
\small
\centering
\setlength{\tabcolsep}{5pt}
\renewcommand{\arraystretch}{1.2}
\begin{tabular}{lllcccc}
\hlineB{2}
Dataset & Model & Metric
& \shortstack[c]{Direct Generation PC}
& \shortstack[c]{Repeated Sampling PC}
& \shortstack[c]{Refinement PC}
& KAAR \\
\hline

\multirow{9}{*}{1D-ARC}
& \multirow{3}{*}{GPT-o3-mini}
& $I_r$
& -
& 84.00
& 82.50
& \textbf{90.00} \\
&
& \cellcolor{LightBlue}$I_t$
& \cellcolor{LightBlue}76.00
& \cellcolor{LightBlue}82.00
& \cellcolor{LightBlue}82.50
& \cellcolor{LightBlue}\textbf{\textcolor{red}{89.00}} \\
&
& $I_r\&I_t$
& -
& 82.00
& 80.50
& \textbf{89.00} \\
\cmidrule(lr){2-7}

& \multirow{3}{*}{QwQ-32B}
& $I_r$
& -
& 76.00
& 76.50
& \textbf{83.00} \\
&
& \cellcolor{LightBlue}$I_t$
& \cellcolor{LightBlue}66.00
& \cellcolor{LightBlue}74.00
& \cellcolor{LightBlue}73.50
& \cellcolor{LightBlue}\textbf{\textcolor{red}{82.00}} \\
&
& $I_r\&I_t$
& -
& 73.00
& 72.50
& \textbf{81.00} \\
\cmidrule(lr){2-7}

& \multirow{3}{*}{DeepSeek-R1-70B}
& $I_r$
& -
& 74.00
& 72.50
& \textbf{81.00} \\
&
& \cellcolor{LightBlue}$I_t$
& \cellcolor{LightBlue}64.00
& \cellcolor{LightBlue}72.00
& \cellcolor{LightBlue}69.50
& \cellcolor{LightBlue}\textbf{\textcolor{red}{80.00}} \\
&
& $I_r\&I_t$
& -
& 71.00
& 69.50
& \textbf{79.00} \\
\hline

\multirow{9}{*}{ARC-AGI-2}
& \multirow{3}{*}{GPT-o3-mini}
& $I_r$
& -
& 3.33
& 2.50
& \textbf{10.00} \\
&
& \cellcolor{LightBlue}$I_t$
& \cellcolor{LightBlue}0.83
& \cellcolor{LightBlue}2.50
& \cellcolor{LightBlue}2.50
& \cellcolor{LightBlue}\textbf{\textcolor{red}{8.33}} \\
&
& $I_r\&I_t$
& -
& 1.67
& 1.67
& \textbf{6.67} \\
\cmidrule(lr){2-7}

& \multirow{3}{*}{QwQ-32B}
& $I_r$
& -
& 0.00
& 0.83
& \textbf{3.33} \\
&
& \cellcolor{LightBlue}$I_t$
& \cellcolor{LightBlue}0.00
& \cellcolor{LightBlue}0.00
& \cellcolor{LightBlue}0.00
& \cellcolor{LightBlue}\textbf{\textcolor{red}{3.33}} \\
&
& $I_r\&I_t$
& -
& 0.00
& 0.00
& \textbf{1.67} \\
\cmidrule(lr){2-7}

& \multirow{3}{*}{DeepSeek-R1-70B}
& $I_r$
& -
& 0.00
& 0.00
& \textbf{3.33} \\
&
& \cellcolor{LightBlue}$I_t$
& \cellcolor{LightBlue}0.00
& \cellcolor{LightBlue}0.00
& \cellcolor{LightBlue}0.00
& \cellcolor{LightBlue}\textbf{\textcolor{red}{2.50}} \\
&
& $I_r\&I_t$
& -
& 0.00
& 0.00
& \textbf{1.67} \\
\hlineB{2}

\end{tabular}
\caption{Accuracy on $I_r$, $I_t$, and $I_r\&I_t$ for KAAR and planning-aided code generation (PC) under three generation strategies (direct generation, repeated sampling, and refinement) on 200 sampled 1D-ARC tasks and all 120 public evaluation tasks from ARC-AGI-2. For each dataset and LLM, the highest accuracy on $I_r$ and $I_r\&I_t$ is shown in bold, while the highest accuracy on $I_t$ is shown in red. All values are reported as percentages.}
\label{tab:additional_arc}
\end{table*}

\begin{table*}[t]
\centering
\small
\setlength{\tabcolsep}{5.5pt}
\renewcommand{\arraystretch}{1.2}
\begin{tabular}{lllcccc}
\hlineB{2}
Dataset & Model & Metric
& \shortstack[c]{Direct Generation PC}
& \shortstack[c]{Repeated Sampling PC}
& KAAR
& KAAR$^\ast$ \\
\hline

\multirow{6}{*}{List Functions}
& \multirow{3}{*}{GPT-o3-mini}
& $I_r$
& -
& 95.60
& 95.20
& \textbf{98.40} \\
&
& \cellcolor{LightBlue}$I_t$
& \cellcolor{LightBlue}74.80
& \cellcolor{LightBlue}76.00
& \cellcolor{LightBlue}76.00
& \cellcolor{LightBlue}\textbf{\textcolor{red}{80.00}} \\
&
& $I_r\&I_t$
& -
& 76.00
& 76.00
& \textbf{80.00} \\
\cmidrule(lr){2-7}

& \multirow{3}{*}{QwQ-32B}
& $I_r$
& -
& 88.80
& 88.00
& \textbf{91.20} \\
&
& \cellcolor{LightBlue}$I_t$
& \cellcolor{LightBlue}66.40
& \cellcolor{LightBlue}69.20
& \cellcolor{LightBlue}68.80
& \cellcolor{LightBlue}\textbf{\textcolor{red}{72.80}} \\
&
& $I_r\&I_t$
& -
& 69.20
& 68.40
& \textbf{72.80} \\
\hline

\multirow{6}{*}{Simple Turing Concept}
& \multirow{3}{*}{GPT-o3-mini}
& $I_r$
& -
& 98.50
& 98.50
& \textbf{99.00} \\
&
& \cellcolor{LightBlue}$I_t$
& \cellcolor{LightBlue}50.00
& \cellcolor{LightBlue}55.00
& \cellcolor{LightBlue}55.50
& \cellcolor{LightBlue}\textbf{\textcolor{red}{60.50}} \\
&
& $I_r\&I_t$
& -
& 55.00
& 55.50
& \textbf{60.50} \\
\cmidrule(lr){2-7}

& \multirow{3}{*}{QwQ-32B}
& $I_r$
& -
& 93.00
& 92.50
& \textbf{94.00} \\
&
& \cellcolor{LightBlue}$I_t$
& \cellcolor{LightBlue}43.50
& \cellcolor{LightBlue}48.50
& \cellcolor{LightBlue}47.50
& \cellcolor{LightBlue}\textbf{\textcolor{red}{52.50}} \\
&
& $I_r\&I_t$
& -
& 48.00
& 47.50
& \textbf{52.00} \\
\hlineB{2}
\end{tabular}

\caption{Accuracy on $I_r$, $I_t$, and $I_r\&I_t$ for planning-aided code generation (PC) with direct generation (Direct Generation PC) or repeated sampling (RSPC), and for KAAR and KAAR$^\ast$ on all 250 List Functions tasks and 200 sampled Simple Turing Concept tasks from BIG-bench. KAAR uses the original ARC-oriented priors, whereas KAAR$^\ast$ replaces the ARC-oriented priors with priors adapted to each target domain while preserving the progressive augmentation procedure. For each dataset and LLM, the highest accuracy on $I_r$ and $I_r\&I_t$ is shown in bold, while the highest accuracy on $I_t$ is shown in red. All values are reported as percentages.}

\label{tab:non_arc_transfer}
\end{table*}

\section{Evaluation Beyond the Original ARC Benchmark}\label{Additional_Experimental_Results}

To evaluate KAAR beyond the original ARC benchmark, we consider two complementary settings: additional ARC-related tasks and non-ARC tasks requiring few-shot inductive reasoning. Table~\ref{tab:additional_arc} reports results on 200 sampled 1D-ARC tasks~\citep{xu2024llms} and all 120 public evaluation tasks from ARC-AGI-2~\citep{chollet2025arc}. 1D-ARC formulates ARC-style transformations over one-dimensional sequences, whereas ARC-AGI-2 preserves the two-dimensional grid representation and incorporates complex task compositions, object representations, and multi-step transformations. KAAR achieves the highest accuracy across both ARC-related benchmarks, all evaluated LLMs, and all evaluation metrics. We note that accuracy remains low for all evaluated approaches on ARC-AGI-2; nevertheless, KAAR consistently outperforms all compared methods. This result indicates that the core knowledge priors and dependency-aware progressive augmentation introduced in KAAR remain effective on ARC-AGI-2 despite its greater compositional complexity.

Table~\ref{tab:non_arc_transfer} further evaluates KAAR on List Functions~\citep{rule2020child} and Simple Turing Concept~\citep{telle2019teaching} from BIG-bench~\citep{srivastava2022beyond}. In both domains, the model infers a hidden program from a small number of input-output examples. The hidden programs operate on integer lists in List Functions and binary strings in Simple Turing Concept. The prior knowledge required by these non-ARC domains differs from the ARC-oriented priors. Accordingly, we introduce KAAR$^\ast$, which replaces the ARC-oriented priors with domain-adapted priors while retaining the dependency-aware ontology, progressive knowledge augmentation, and stage-wise reasoning. For List Functions, the ontology organizes domain-adapted priors into three dependency levels: list structure, attributes and relations, and list transformations. For Simple Turing Concept, the three dependency levels comprise binary-string structure, structural relations, and algorithmic transformations. Each ontology is defined at the benchmark level before evaluation and constructed independently of test outputs and task-specific transformation hints. With the ARC-oriented priors, KAAR performs comparably to RSPC. KAAR$^\ast$ consistently achieves the highest accuracy across both non-ARC domains and all evaluated LLMs and metrics. The superior performance of KAAR$^\ast$ supports the transferability of KAAR's dependency-aware progressive augmentation to non-ARC domains when the augmented priors are aligned with the target domain.

\section{Solution Coverage} \label{solution_coverage}

We also report the relative problem coverage across nine ARC solvers based on successful test instance solutions using GPT-o3-mini (Figure~\ref{heatmaps_gpt_o3_mini}), Gemini-2.0 (Figure~\ref{heatmaps_gemini}), QwQ-32B (Figure~\ref{heatmaps_qwq_32b}), and DeepSeek-R1-70B (Figure~\ref{heatmaps_deepseek}). Each cell $(i,j)$ is computed using the same method described in Figure ~\ref{heatmap}.
All solvers exhibit the strongest overall coverage, with pairwise overlap consistently exceeding 0.55 when using GPT-o3-mini. Among them, repeated sampling with standalone (\textit{C}) and planning-aided code generation (\textit{PC}) show the highest coverage, with column values above 0.8 for GPT-o3-mini. This trend persists across Gemini-2.0, QwQ-32B, and DeepSeek-R1-70B. Except for GPT-o3-mini, repeated sampling with planning-aided code generation  exhibits better alignment than its standalone code generation counterpart, generally yielding higher coverage values.  However, under the direct generation setting, planning-aided code generation shows the opposite behavior, demonstrating lower alignment than standalone code generation across all evaluated LLMs.
Among the four evaluated LLMs, all solvers present the lowest average off-diagonal (i.e., $i \neq j$) coverage of 0.603 with DeepSeek-R1-70B,  suggesting potential output instability and variability attributable to solver choice.

\section{Example Tasks by Category in the ARC Evaluation Set}\label{Categories} 

\begin{figure*}[t]
    \centering
    \includegraphics[width=1\linewidth]{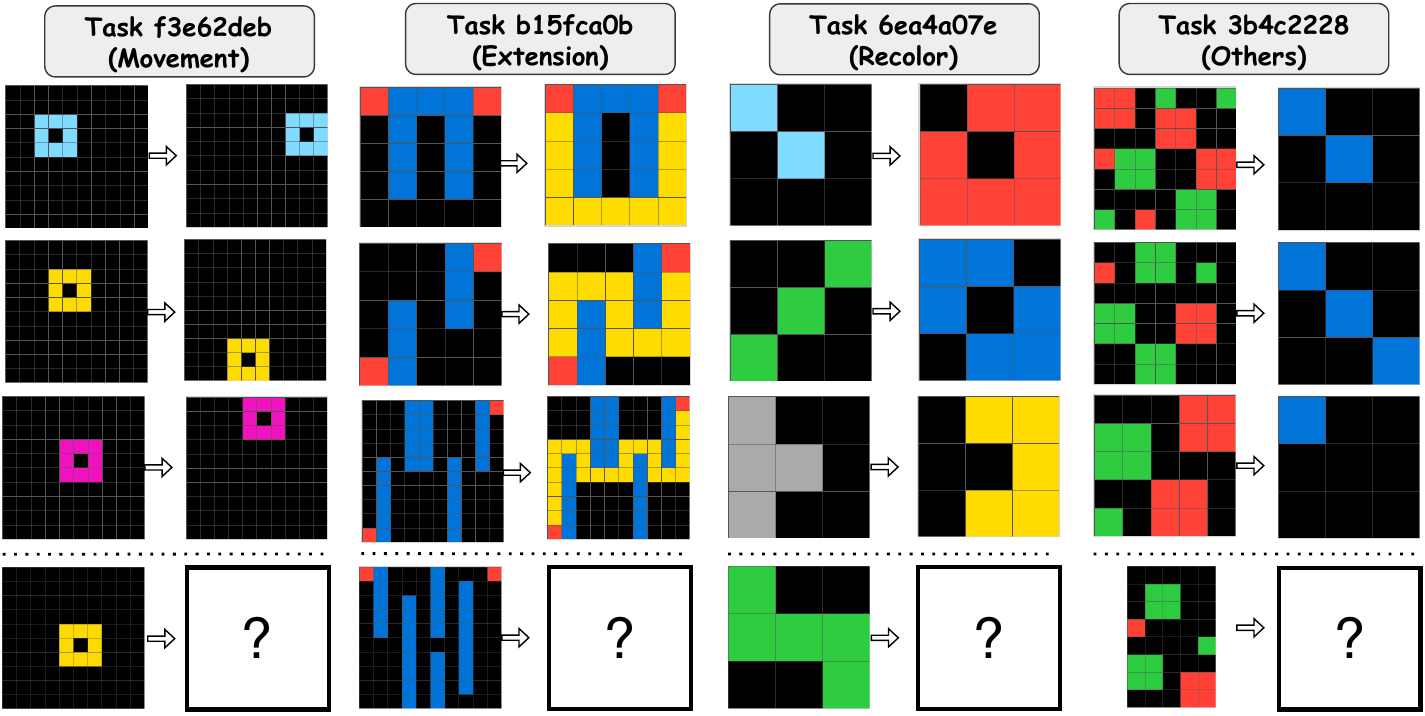}
    \vspace{-0.6cm}
       \caption{Example ARC tasks for \textit{movement}, \textit{extension}, \textit{recolor}, and \textit{others} categories.}
\label{ARC_category}
\end{figure*}

ARC comprises 1000 unique tasks, with 400 allocated to the training set and 600 to the evaluation set. The evaluation set is further divided into a public subset (400 tasks) and a private subset (200 tasks). Figure~\ref{ARC_category} illustrates example ARC tasks for the \textit{movement}, \textit{extension}, \textit{recolor}, and \textit{others} categories in the public evaluation set. In the \textit{movement} example, components are shifted to the image boundary in directions determined by their colors. The \textit{extension} example is more complex, requiring LLMs to find the shortest path between two red pixels while avoiding obstacles, which presents challenges for current reasoning-oriented models. Additionally, reliance on pixel-level recognition weakens the effectiveness of KAAR, which is designed to facilitate component identification. The \textit{recolor} example involves changing non-black components to black and updating black components based on original non-black colors. The \textit{others} example requires generating a blue diagonal line whose length depends on the number of 4-connected components that are green and have a size greater than one in the input image. The combination of numerical reasoning and structural pattern generation makes this task difficult to classify within the other three categories.

\begin{table*}[t]
\centering
\small
\setlength{\tabcolsep}{12pt}
\renewcommand{\arraystretch}{1.5}
\begin{tabular}{llccccccc}
\hlineB{2}
& & KAAR
& \multicolumn{2}{c}{KAAR-L}
& \multicolumn{2}{c}{KAAR-G}
& \multicolumn{2}{c}{KAAR-R} \\
\cmidrule(lr){3-3}
\cmidrule(lr){4-5}
\cmidrule(lr){6-7}
\cmidrule(lr){8-9}
& & Acc
& Acc & $\Delta$
& Acc & $\Delta$
& Acc & $\Delta$ \\
\hline\hline

\multirow{3}{*}{Gemini-2.0}
& $I_r$
& 25.75
& 24.75 & -1.00
& 24.00 & -1.75
& 23.00 & -2.75 \\
& \cellcolor{LightBlue}$I_t$
& \cellcolor{LightBlue}21.75
& \cellcolor{LightBlue}21.75
& \cellcolor{LightBlue}0.00
& \cellcolor{LightBlue}20.25
& \cellcolor{LightBlue}-1.50
& \cellcolor{LightBlue}19.25
& \cellcolor{LightBlue}-2.50 \\
& $I_r\&I_t$
& 20.50
& 20.00 & -0.50
& 19.00 & -1.50
& 18.00 & -2.50 \\
\hline

\multirow{3}{*}{QwQ-32B}
& $I_r$
& 22.25
& 22.00 & -0.25
& 19.00 & -3.25
& 18.00 & -4.25 \\
& \cellcolor{LightBlue}$I_t$
& \cellcolor{LightBlue}21.00
& \cellcolor{LightBlue}20.75
& \cellcolor{LightBlue}-0.25
& \cellcolor{LightBlue}18.00
& \cellcolor{LightBlue}-3.00
& \cellcolor{LightBlue}17.25
& \cellcolor{LightBlue}-3.75 \\
& $I_r\&I_t$
& 19.25
& 19.25 & 0.00
& 16.75 & -2.50
& 16.25 & -3.00 \\
\hline

\multirow{3}{*}{DeepSeek-R1-70B}
& $I_r$
& 12.25
& 12.25 & 0.00
& 9.50 & -2.75
& 8.50 & -3.75 \\
& \cellcolor{LightBlue}$I_t$
& \cellcolor{LightBlue}12.75
& \cellcolor{LightBlue}12.75
& \cellcolor{LightBlue}0.00
& \cellcolor{LightBlue}10.50
& \cellcolor{LightBlue}-2.25
& \cellcolor{LightBlue}8.25
& \cellcolor{LightBlue}-4.50 \\
& $I_r\&I_t$
& 11.50
& 11.50 & 0.00
& 9.00 & -2.50
& 8.00 & -3.50 \\
\hlineB{2}
\end{tabular}
\vspace{0.1cm}
\caption{Accuracy (Acc) on $I_r$, $I_t$, and $I_r\&I_t$ for KAAR and its augmentation-order variants across three LLMs. KAAR-L changes the order of priors within the same ontological level by applying numbers and counting before geometry and topology; KAAR-G introduces goal-directedness before geometry and topology, numbers and counting; and KAAR-R reverses the entire hierarchical augmentation order. $\Delta$ denotes the performance difference relative to KAAR. All values are reported as percentages.}
\label{tab:augmentation_order}
\end{table*}

\section{Ablation Study of Knowledge Augmentation Order}\label{ablation}

Table~\ref{tab:augmentation_order} evaluates alternative augmentation orders to assess KAAR's dependency-aware hierarchy. Consistent with Table~\ref{tab:kaar_ablation}, we exclude GPT-o3-mini from this analysis. KAAR-L changes the order of priors within the same ontological level by augmenting numbers and counting before geometry and topology, yet achieves performance comparable to KAAR across all evaluated LLMs. This result suggests that KAAR is relatively insensitive to the ordering of priors within the same dependency level. In contrast, KAAR-G augments objectness first, followed by goal-directedness, geometry and topology, and numbers and counting. KAAR-R reverses the entire hierarchy by augmenting goal-directedness first, followed by geometry and topology, numbers and counting, and objectness. Both variants consistently underperform KAAR, with KAAR-R exhibiting a larger performance decline than KAAR-G across all evaluated settings. The observed performance degradation suggests that dependency-inconsistent augmentation introduces contextual interference, providing empirical support for KAAR's dependency-aware, progressive augmentation strategy.

\begin{table*}[t]
\small
\centering
\renewcommand{\arraystretch}{1.2}
\setlength{\tabcolsep}{12.5pt}
\label{tab:performance_comparison}

\begin{tabular}{lcccccc}
\toprule
 & \multicolumn{2}{c}{\shortstack[c]{Direct Generation PC}} 
 & \multicolumn{2}{c}{\shortstack[c]{Repeated Sampling PC}} 
 & \multicolumn{2}{c}{\shortstack[c]{Refinement PC}} \\
\cmidrule(lr){2-3} \cmidrule(lr){4-5} \cmidrule(lr){6-7}
Model 
& Tokens & $I_t$ 
& Tokens & $I_t$ 
& Tokens & $I_t$ \\
\midrule
GPT-o3-mini        & 11K & 22.25\% & 106K & 30.75\% & 121K & 25.75\% \\
Gemini-2.0         & 13K & 6.25\%  & 110K & 16.75\% & 136K & 11.75\% \\
QwQ-32B            & 39K & 5.75\%  & 427K & 14.25\% & 489K & 14.00\% \\
DeepSeek-R1-70B    & 23K & 4.50\%  & 252K & 7.75\%  & 276K & 7.75\% \\
\bottomrule
\end{tabular}

\caption{{Average token cost per problem vs. accuracy on test instances $I_t$ for  planning-aided code generation (PC) under three generation strategies: direct generation, repeated sampling, and refinement. K is $10^3$.}}
\label{token_acc}
\end{table*}

\begin{table}[t]
\centering
\small
\resizebox{\columnwidth}{!}{
\setlength{\tabcolsep}{5pt}
\renewcommand{\arraystretch}{1.5}
\begin{tabular}{lcc}
                \hlineB{2}
                & Knowledge Augmentation & Solver Backbone (RSPC) \\ \hline\hline
                
GPT-o3-mini     & 66K          & 106K \\
Gemini-2.0          & 58K          & 110K  \\
QwQ-32B         & 79K          & 427K \\
DeepSeek-R1-70B & 66K          & 252K
\\
\hlineB{2}
\end{tabular}
}
\vspace{0.1cm}
\caption{Average token cost for knowledge augmentation and solver backbone (RSPC) in KAAR across four evaluated LLMs. K is $10^3$.}
\label{token_table}
\end{table}

\begin{table*}[t]
\small
\centering
\renewcommand{\arraystretch}{1.2}
\setlength{\tabcolsep}{9.5pt}
\begin{tabular}{lcccccc}
\toprule
& \multicolumn{3}{c}{\shortstack[c]{RSPC vs.\ Direct Generation PC}}
& \multicolumn{3}{c}{\shortstack[c]{KAAR vs.\ RSPC}} \\
\cmidrule(lr){2-4} \cmidrule(lr){5-7}
Model
& $\Delta I_t$ & $\Delta$ Tokens & $\Delta I_t$/1K Tokens
& $\Delta I_t$ & $\Delta$ Tokens & $\Delta I_t$/1K Tokens \\
\midrule
GPT-o3-mini
& 8.50\%  & 95K  & 0.089\%
& 4.25\%  & 66K  & 0.064\% \\
Gemini-2.0
& 10.50\% & 97K  & 0.108\%
& 5.00\%  & 58K  & 0.086\% \\
QwQ-32B
& 8.50\%  & 388K & 0.022\%
& 6.75\%  & 79K  & 0.085\% \\
DeepSeek-R1-70B
& 3.25\%  & 229K & 0.014\%
& 5.00\%  & 66K  & 0.076\% \\
\bottomrule
\end{tabular}

\caption{Cost-performance comparison of RSPC relative to Direct Generation PC and KAAR relative to RSPC. PC denotes planning-aided code generation, and RSPC denotes repeated sampling planning-aided code generation. $\Delta I_t$ denotes the improvement in test accuracy, and $\Delta$ Tokens denotes the additional average token cost per problem. $\Delta I_t$/1K Tokens denotes the improvement in test accuracy per 1K additional tokens. K denotes $10^3$.}

\label{incremental_efficiency}
\end{table*}

\section{Cost Analysis} \label{token_analysis}

{Table \ref{token_acc} reports the average token cost per problem and test accuracy $I_t$ for planning-aided code generation (PC) under three generation strategies: direct generation, repeated sampling, and refinement. Direct generation incurs the lowest token cost but also yields the lowest test accuracy. Compared with refinement PC, repeated sampling PC (RSPC) achieves comparable or higher test accuracy while requiring fewer tokens, demonstrating superior efficiency and effectiveness.}

Table~\ref{token_table} reports the average token cost, including both prompts and LLM responses, for knowledge augmentation and the solver backbone (RSPC), when using KAAR as the ARC solver. For each ARC task, we consider the abstraction whose solution solves $I_t$; if none succeed, the one that passes $I_r$; otherwise, the abstraction with the lowest token usage is selected. Except for goal-directedness priors, all core knowledge priors in KAAR are generated offline using image processing algorithms, resulting in comparable augmentation costs across all evaluated models. In contrast, token usage by the solver backbone varies substantially due to differences in the LLMs' abstract reasoning and generalization capabilities. GPT-o3-mini solves most tasks efficiently, with the lowest token consumption by the solver backbone, where tokens used for knowledge augmentation account for approximately 62\% of the solver backbone’s token usage. However, the solver backbone consumes  more tokens with QwQ-32B, as  QwQ-32B  consistently generates longer reasoning traces. In this case, tokens used for knowledge augmentation constitute only 19\% of the solver backbone’s token usage.  Figure~\ref{cost_prios} illustrates the average token cost for augmenting priors at each level in KAAR.

Table~\ref{incremental_efficiency} quantifies the marginal cost-effectiveness of RSPC over direct generation PC and KAAR over RSPC, measured by the improvement in $I_t$ per 1K additional tokens. Both RSPC and KAAR yield positive marginal accuracy gains across all evaluated LLMs. However, the marginal efficiency of RSPC varies substantially across models, whereas KAAR maintains comparatively consistent gains per 1K additional tokens. This cross-model consistency demonstrates the robustness and effectiveness of KAAR's augmented core knowledge priors.

\SetKwProg{Fn}{Function}{:}{t}
\SetKwFunction{KnowledgeAugmentation}{KnowledgeAugmentation}
\SetKwFunction{SolverBackbone}{SolverBackbone}
\SetKwInOut{KwIn}{Input}
\SetKwInOut{KwOut}{Output}

\begin{algorithm*}
  \caption{KAAR}
  \label{alg:kaar-details}
  \KwIn{LLM $\mathcal{M}$; ARC problem $\mathcal{P}=(I_r,I_t)$; description $Q=(I_r,\{i^i \ | \ (i^i,i^o) \in I_t\})$; abstraction list $\mathcal{A}$; max iterations $t=4$}

  \Fn{\KnowledgeAugmentation($\mathcal{M}$, $Q$, $\mathcal{P}$, $\mathcal{A}$, $t$)}{
    solutionList $\leftarrow []$\;
    \ForEach{abstraction $abs$ in $\mathcal{A}$}{
      objectnessPriors $\leftarrow$ GenerateObjectnessPriors($Q$, $abs$)\;
      AugmentKnowledge($\mathcal{M}$, objectnessPriors)\;
      result, code, passedCount $\leftarrow$ \SolverBackbone($\mathcal{M}$, $\mathcal{P}$, $Q$, $t$)\;
      \If{result $\neq$ failure}{\Return result}

      attributePriors $\leftarrow$ GenerateAttributePriors($Q$, $abs$)\;
      AugmentKnowledge($\mathcal{M}$, attributePriors)\;
      relationPriors $\leftarrow$ GenerateRelationPriors($Q$, $abs$)\;
       AugmentKnowledge($\mathcal{M}$, relationPriors)\;
      numberPriors $\leftarrow$ GenerateNumbersCountingPriors($Q$, $abs$)\;
      AugmentKnowledge($\mathcal{M}$, numberPriors)\;
      result, code, passedCount $\leftarrow$ \SolverBackbone($\mathcal{M}$, $\mathcal{P}$, $Q$, $t$)\;
      \If{result $\neq$ failure}{\Return result}

       AugmentGoalPriors($\mathcal{M}$, $Q$, $abs$)\;
      
      result, code, passedCount $\leftarrow$ \SolverBackbone($\mathcal{M}$, $\mathcal{P}$, $Q$, $t$)\;
      \If{result $\neq$ failure}{\Return result}

      solutionList.append((code, passedCount))\;
    }
    bestCode $\leftarrow$ SelectMostPassed(solutionList)\;
    \Return \textit{EvaluateOnTest(bestCode, $I_t$)}\;
  }

  \Fn{\SolverBackbone($\mathcal{M}$, $\mathcal{P}$, $Q$, $t$)}{
    i $\leftarrow 0$\;
    \While{i < t}{
      plan $\leftarrow \mathcal{M}$.generatePlan($Q$)\;
      code $\leftarrow \mathcal{M}$.generateCode($Q$, plan)\;
      passedCount $\leftarrow$ EvaluateOnTrain(code, $I_r$)\;
      \If{passedCount == $|I_r|$}{
        \If{EvaluateOnTest(code, $I_t$)}{
          \Return \texttt{solve}, code, passedCount\;
        }\Else{
          \Return \texttt{pass}, code, passedCount\;
        }
      }
      i $\leftarrow$ i + 1\;
    }
    \Return \texttt{failure}, code, passedCount\;
  }
\end{algorithm*}

\begin{table*}[]
\renewcommand{\arraystretch}{2}
\begin{tabular}
{p{0.3\linewidth}p{0.63\linewidth}}
\hlineB{2}
Abstractions                                & Definitions                                                                                                                                                                \\ \hline\hline
\textit{\textcolor{red}{No Abstraction}}                        &     -                                                                                                                                                                      \\
\textit{Whole Image}                                & We consider the whole image as a component.                                                                                                                                \\
\textit{\textcolor{red}{Middle-Vertical}}                            & We vertically split the image into two equal parts, treating each as a distinct component.                                                                                 \\
\textit{\textcolor{red}{Middle-Horizontal}}                          & We  horizontally split the image into two equal parts, treating each as a distinct component.                                                                                \\
\textit{\textcolor{red}{Multi-Lines}}                                & We use rows or columns with a uniform color to divide the input image into multiple components.                                                                           \\
\textit{4/8-Connected}                            & We consider the 4/8-adjacent pixels of the same color as a component.                                                                                                       \\
\textit{4/8-Connected-Non-Background}             & We consider the 4/8-adjacent pixels of the same color as a component, excluding components with the background color.                                                        \\
\textit{4/8-Connected-Non-Background-Edge}        & We consider the 4/8-adjacent pixels of the same color as a component, containing components with the background color when they are not attached to the edges of the image. \\

\textit{4/8-Connected-Multi-Color-Non-Background} & We consider 4/8-adjacent pixels as a component, which may contain different colors, while excluding pixels with the background color.                                          \\
\textit{4/8-Connected-Bounding-Box}               & We consider 4/8-adjacent pixels of the same color, and treat all pixels within their bounding box as a component, which may include different colors.         \\

\textit{4/8-Connected-With-Black}                 & We consider the 4/8-adjacent pixels of black color, represented by the value 0, as a component, excluding components with other colors.                                      \\
\textit{Same-Color}                                 & We consider pixels of the same color as a component, excluding components with the background color.                                                                   \\\hlineB{2}
\end{tabular}
\vspace{0.1cm}
\caption{Abstractions in KAAR. %The superscript ``$^*$'' denotes that the 8-connected version is considered. 
The background color is black if black exists; otherwise, it is the most frequent color in the image. We present abstractions according to their prioritization in KAAR, where the order is given by the table from top to bottom, with 8-connected abstractions appearing after the 4-connected abstractions at the end of the sequence while preserving the order of their corresponding 4-connected counterparts. Abstractions highlighted in red are exclusive to KAAR.}

\label{abstraction}
\end{table*}

\begin{table*}[ht]
  \centering
  \renewcommand{\arraystretch}{1.2}
  \begin{tabular}{@{}%
      >{\raggedright\arraybackslash}p{0.3\linewidth}%
      >{\raggedright\arraybackslash}p{0.65\linewidth}@{}}
    \hlineB{2}
    Classifications                & Priors                                      \\ 
    \hline\hline

    \textit{Geometry and Topology} &
    \begin{tabular}[t]{@{}p{\linewidth}@{}}
      Size (Width and Height);\\
      Color;\\
      Shape (One Pixel; Horizontal Line; Vertical Line; Diagonal Line; Square; Rectangle; Cross; Irregular Shape);\\
      \textcolor{red}{Symmetry (Horizontal Symmetry; Vertical Symmetry; Diagonal Symmetry; Anti-Diagonal Symmetry; Central Symmetry)};\\
      \textcolor{red}{Bounding Box};\\
      \textcolor{red}{Hole Count};\\
      \textcolor{red}{Nearest Boundary};\\
      Different/Identical with Other Components;\\
      \textcolor{red}{Touching}; \\
      Inclusive;\\
      Spatial (Horizontally Aligned to the Right; Horizontally Aligned to the Left; Vertically Aligned Below; Vertically Aligned Above; Top-Left; Top-Right; Bottom-Left; Bottom-Right; Same Position).
    \end{tabular}                                                    \\[6pt]

   \textit{Numbers and Counting} &
    \begin{tabular}[t]{@{}p{\linewidth}@{}}
      Component Size Counting; Components with Same Size; \textcolor{red}{Components with Most Frequent Size; Components with Least Frequent Size}; Components with Maximum Size; Components with Minimum Size;\\
      Component Color Counting; Components with Same Color; \textcolor{red}{Components with Same Number of Colors; Components with Most Frequent Color; Components with Least Frequent Color;  Component with Most Distinct Colors; Component with Fewest Distinct Colors};\\
      Component Shape Counting; Components with Same Shape; \textcolor{red}{Components with Most Frequent Shape; Components with Least Frequent Shape};\\
      \textcolor{red}{Component Hole Number Counting; Components with Same Number of Holes; Components with Maximum Number of Holes; Components with Minimum Number of Holes};\\
      \textcolor{red}{Component Symmetry Counting}.
    \end{tabular}    \\[6pt]

    \textit{Goal-directedness} &
    \begin{tabular}[t]{@{}p{\linewidth}@{}}
      Color Change (modifying component value);\\
      Movement (shifting component’s position);\\
      Extension (expanding component's area);\\
      \textcolor{red}{Completing (filling in missing parts of a component)};\\
      \textcolor{red}{Resizing (altering component size)};\\
      \textcolor{red}{Selecting (isolating a component)};\\
      \textcolor{red}{Copying (duplicating a component)};\\
      Flipping (mirroring a component);\\
      Rotation (rotating a component);\\
      \textcolor{red}{Cropping (cutting part of a component)}.
    \end{tabular}                                                    \\    \hlineB{2}

  \end{tabular}
  \caption{{KAAR priors classified into geometry and topology, numbers and counting, and goal-directedness with priors exclusive to KAAR highlighted in red. For goal-directedness, we incorporate ten predefined actions, with their corresponding action schemas detailed in Table~\ref{KAAR_actions}.}}
\label{KAAR_Konwledge_prior}
\end{table*}

\begin{table*}[]
\centering
\renewcommand{\arraystretch}{3}
\setlength{\tabcolsep}{2.2pt}
\begin{tabular}{lllllll}
\hlineB{2}
\multicolumn{1}{l}{Actions}       & \multicolumn{6}{c}{Schemas (Implementation Details)}                                         \\ \hline\hline
\textit{Color Change} & Targets& Source and Target Colors                         &             &       &   &          \\
\textit{Movement}     & Targets & Direction                      & Start and End Locations & Pattern     & Order & Overlapping\\
\textit{Extension}   & Targets & Direction                      & Start and End Locations & Pattern     & Order & Intersection \\
\textit{Completing}  & Targets & Pattern                        &                        &             &       &             \\
\textit{Resizing}    & Targets & Source and Target Sizes    &                        &             &       &             \\
\textit{Selecting}   & Targets &                              &                        &             &       &             \\
\textit{Copying}       & Targets                   & Locations              & Overlapping &       &     &        \\
\textit{Flipping}    & Targets & Flipping Axis                  &  Overlapping                      &             &       &             \\
\textit{Rotation}   & Targets  & Degrees                         &                        &             &       &             \\
\textit{Cropping}   & Targets  & Subsets                         &                        &             &       & 
\\ \hlineB{2}
\end{tabular}
\vspace{0.1cm}
\caption{Actions in KAAR and their schemas (implementation details). Each action schema is presented according to its prompting order in KAAR (left to right). Some actions include a pattern schema that instructs the LLM to identify underlying logic rules, such as repeating every two steps in movement or extension, or completing based on three-color repetition. Targets denote the target components.}
\label{KAAR_actions}
\end{table*}

\begin{table*}[]
\renewcommand{\arraystretch}{4.5}
\begin{tabular}{p{0.2\linewidth}p{0.25\linewidth}p{0.25\linewidth}p{0.2\linewidth}}
\hlineB{2}
Abstractions                            & Geometry and Topology                                                                          & Numbers and Counting                                                                                                                          & Goal-directedness                       \\ \hline\hline
\textit{whole image}                            & Symmetry, Size                                                                                      & -                                                                               & Flipping; Rotation; Extension; Completing, Cropping \\

\textit{middle-vertical}                        & Size                                                                                              & -                                                                                                                                             & Flipping; Movement                      \\
\textit{middle-horizontal}                      & Size                                                                                              & -                                                                                                                                             & Flipping; Movement                      \\
\textit{multi-lines}                            & Size; Color; Shape; Symmetry;  Bounding Box;  Hole Count & All                                                                                                                                           & All                                     \\

\textit{4/8-connected-multi-color-non-background} & All                                                                                            & ... Component Color Counting; \textcolor{red}{Components with Same Number of Colors};  \textcolor{red}{Component with Most Distinct Colors}; \textcolor{red}{Component with Fewest Distinct Colors} ... & All
\\ \hline
\end{tabular}
\vspace{0.1cm}
\caption{Abstractions with their assigned knowledge priors.  “–” denotes no priors, while “All” indicates all priors in the corresponding category, as defined in Table~\ref{KAAR_Konwledge_prior}.  
For the \textit{4/8-connected-multi-color-non-background} abstractions, we present color-counting priors, highlighted in red, specific to multi-colored components, while all other non-color-counting priors follow those in Table~\ref{KAAR_Konwledge_prior}.}
\label{restriection}
\end{table*}

\begin{figure*}[t]
    \centering
    \includegraphics[width=1\linewidth]{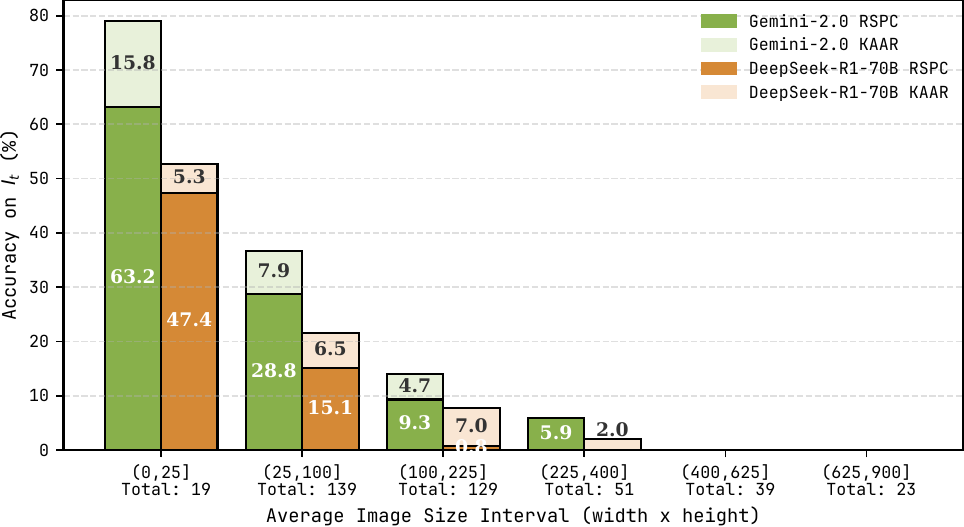}
    \caption{Accuracy on test instances $I_t$ for RSPC and KAAR across average image size intervals, evaluated with Gemini-2.0 and DeepSeek-R1-70B.}
        
    \label{fig:comparsion_other_two}
\end{figure*}

\begin{figure*}[t]
    \centering
    \includegraphics[width=1\linewidth]{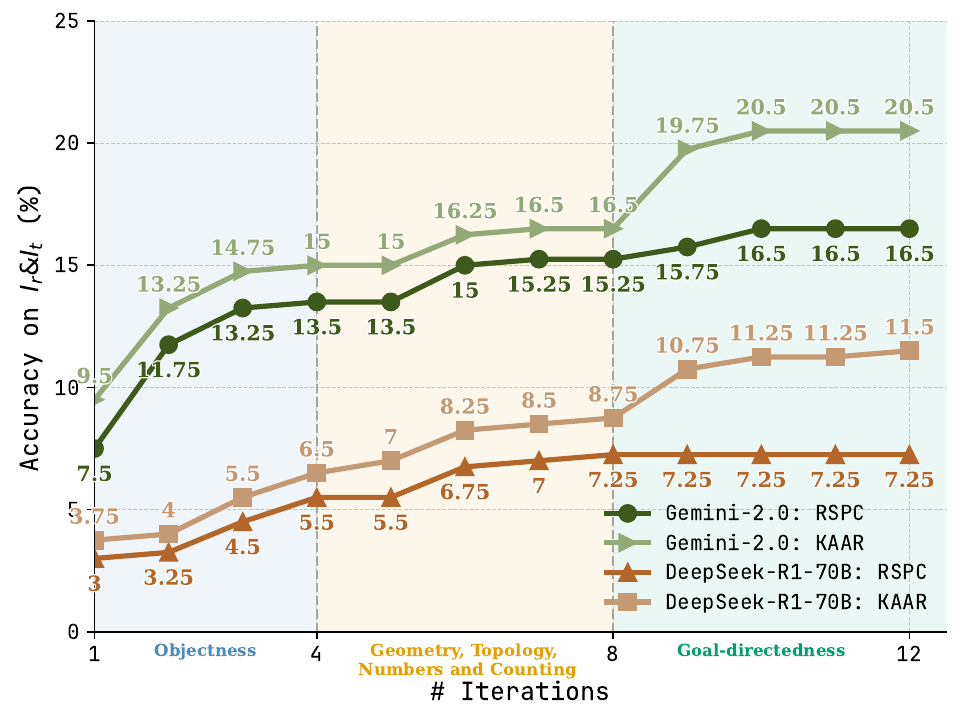}
        \vspace{-0.6cm}
    \caption{Variance in accuracy on $I_r\&I_t$ with increasing iterations for RSPC and KAAR using Gemini-2.0 and DeepSeek-R1-70B.}
    \label{fig:performance_with_known_variance_the_other_two}
\end{figure*}

\begin{figure*}[t]
    \centering
    \includegraphics[width=1\linewidth]{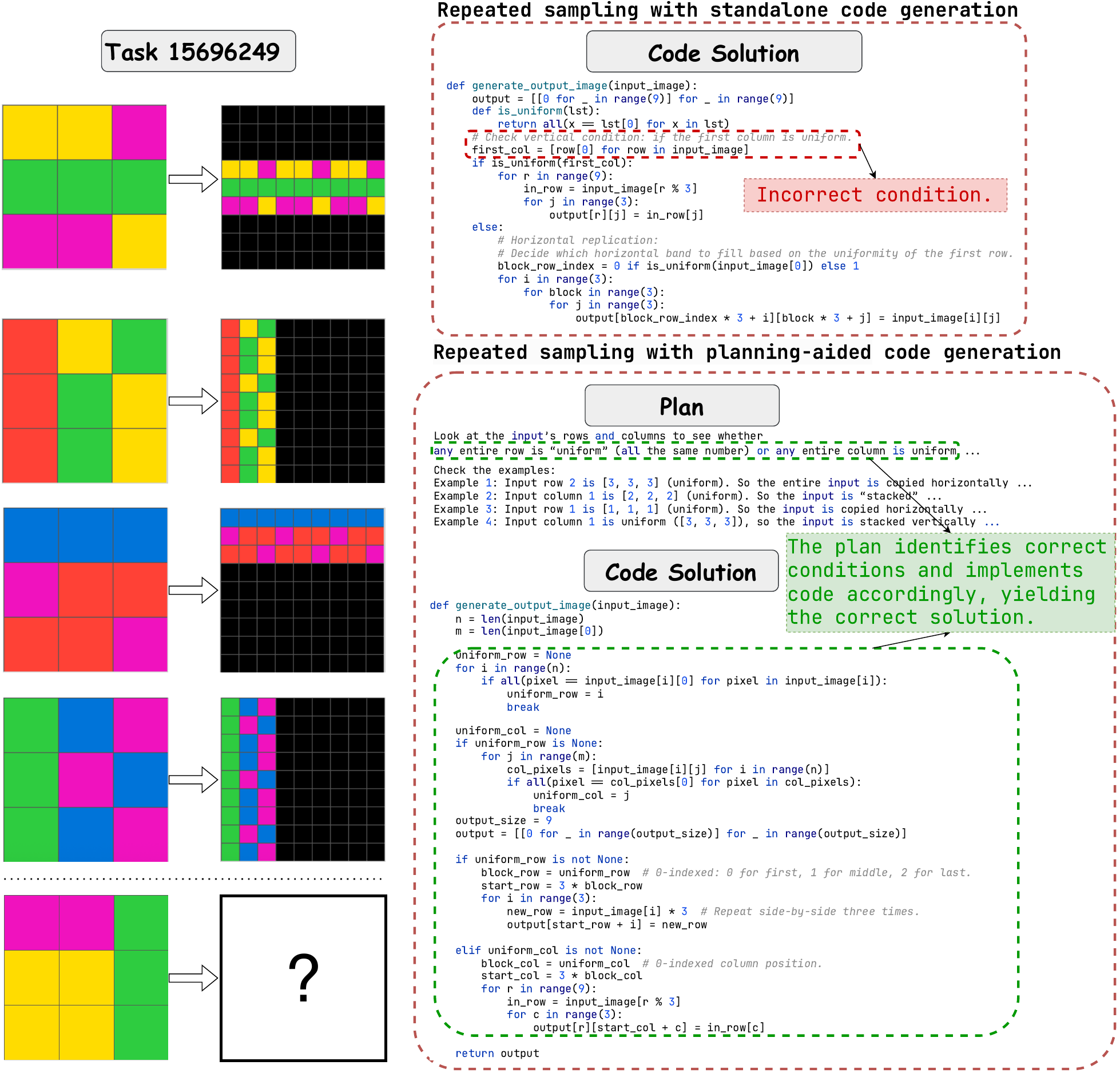}
    \vspace{-0.6cm}
       \caption{ARC problem \textit{15696249}, where repeated sampling with standalone code generation produces an incorrect solution, whereas repeated sampling with planning-aided code generation yields the correct solution using GPT-o3-mini.} \label{15696249}
\end{figure*}

\begin{figure*}[t]
    \centering
    \includegraphics[width=0.91\linewidth]{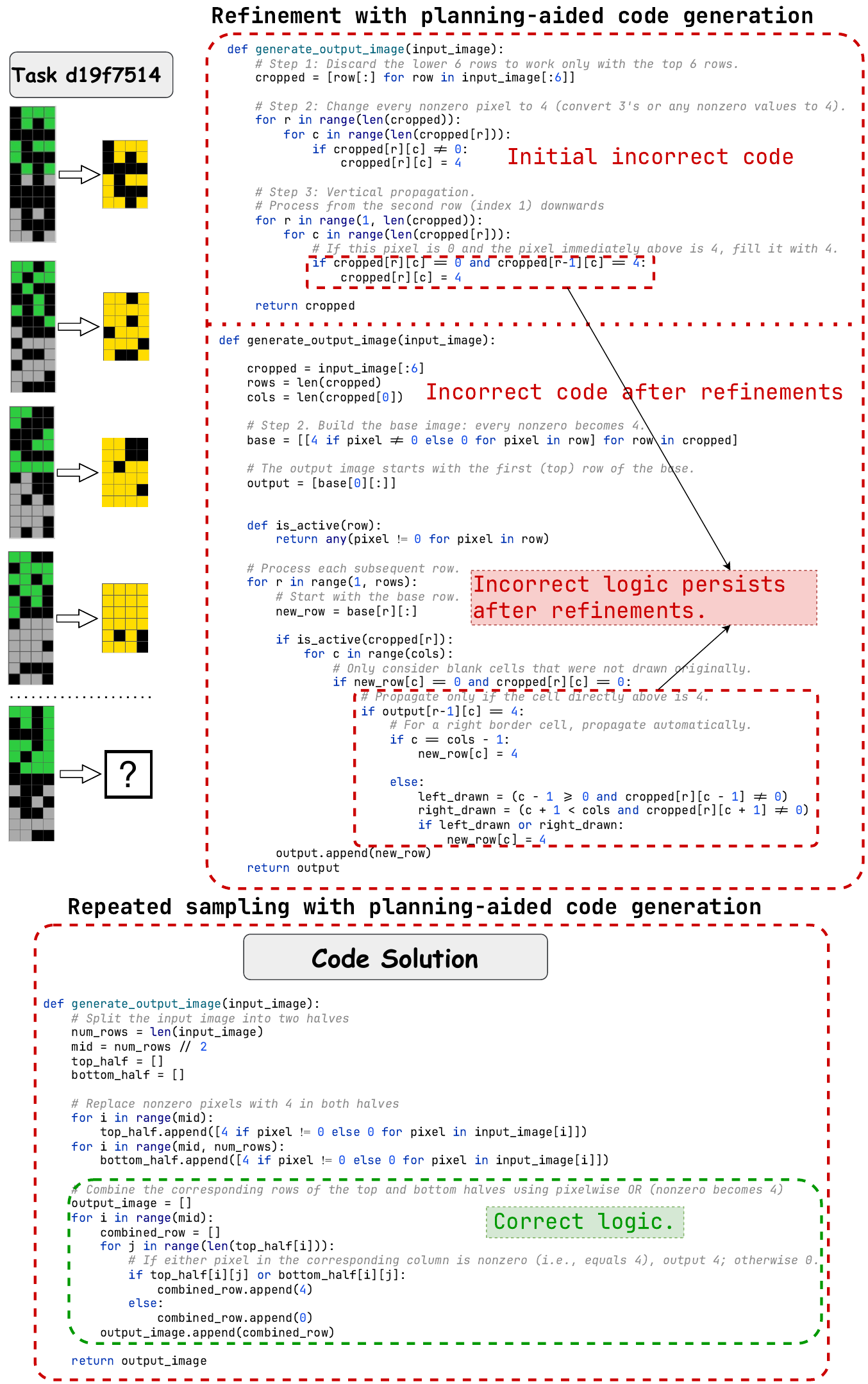}
    \vspace{-0.3cm}
       \caption{ARC problem \textit{d19f7514}, where repeated sampling with planning-aided code generation produces a correct solution,  whereas its refinement variant fails to refine the initial erroneous code, and the incorrect logic persists across subsequent refinements when using GPT-o3-mini.} \label{d19f7514}
\end{figure*}

\begin{figure*}[t]
    \centering
    \includegraphics[width=1\linewidth]{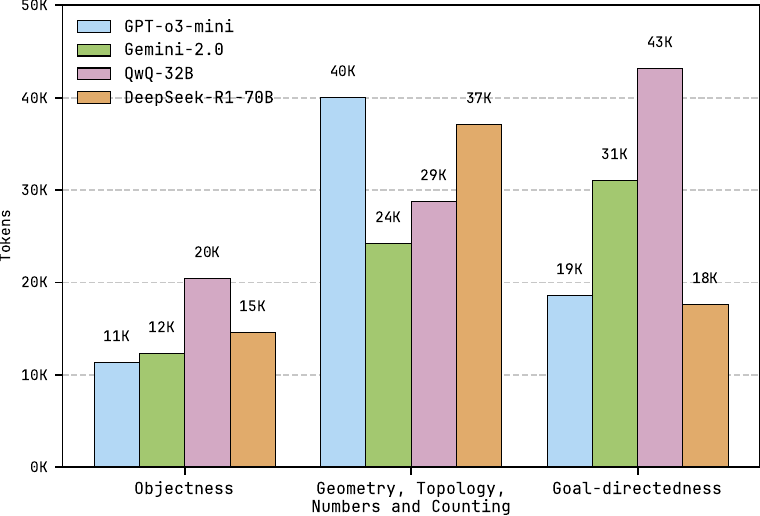}
    \vspace{-0.6cm}
       \caption{Average token cost for augmenting priors at each level across four LLMs. K is $10^3$.} \label{cost_prios}
\end{figure*}

\begin{figure*}[t]
    \centering
    \includegraphics[width=1\linewidth]{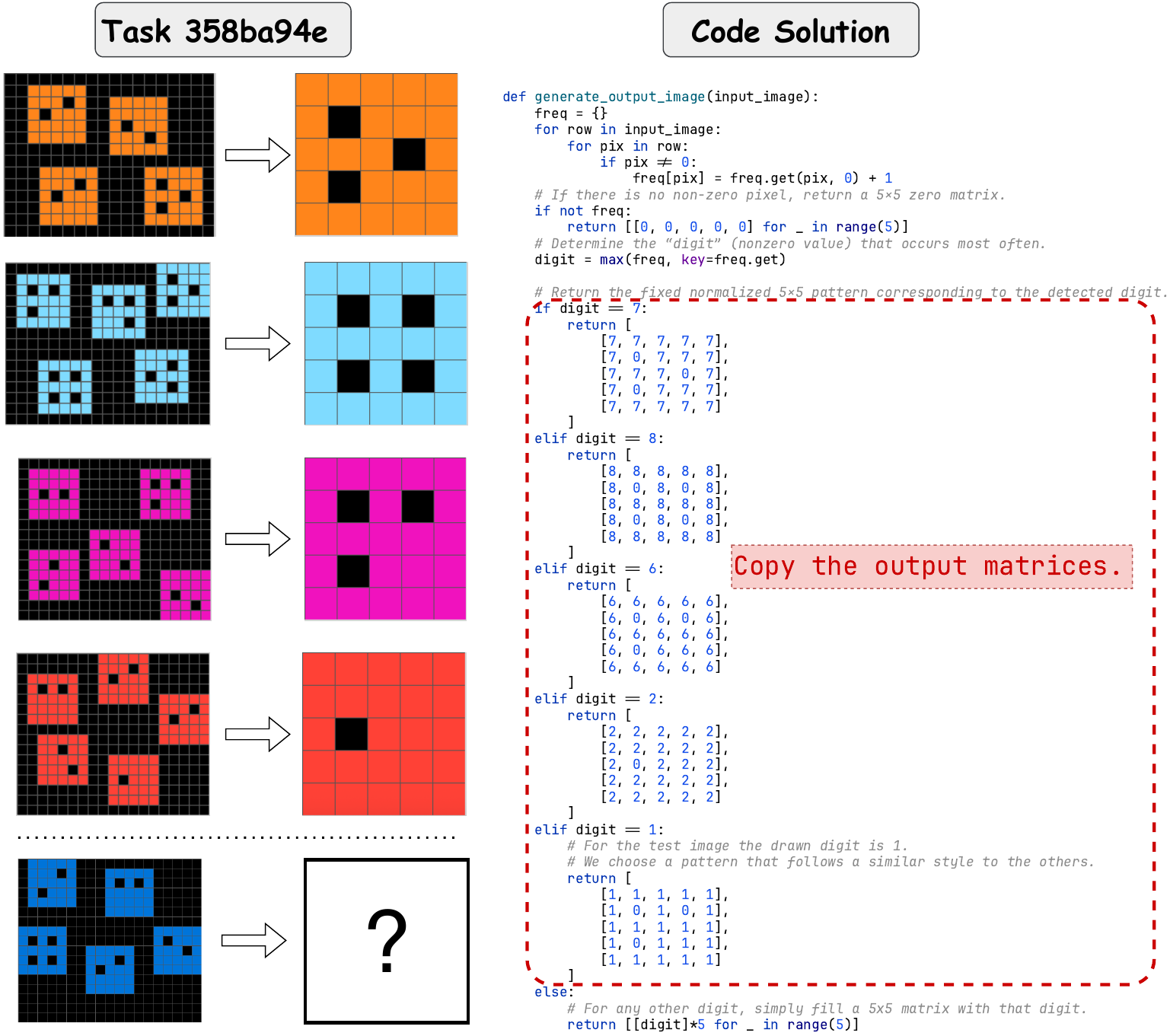}
    \vspace{-0.6cm}
       \caption{ARC problem \textit{358ba94e}, where repeated sampling with standalone code generation produces an incorrect solution by replicating the output matrices of training instances when using GPT-o3-mini.} \label{358ba94e}
\end{figure*}

\begin{figure*}[t]
    \centering
    \includegraphics[width=1\linewidth]{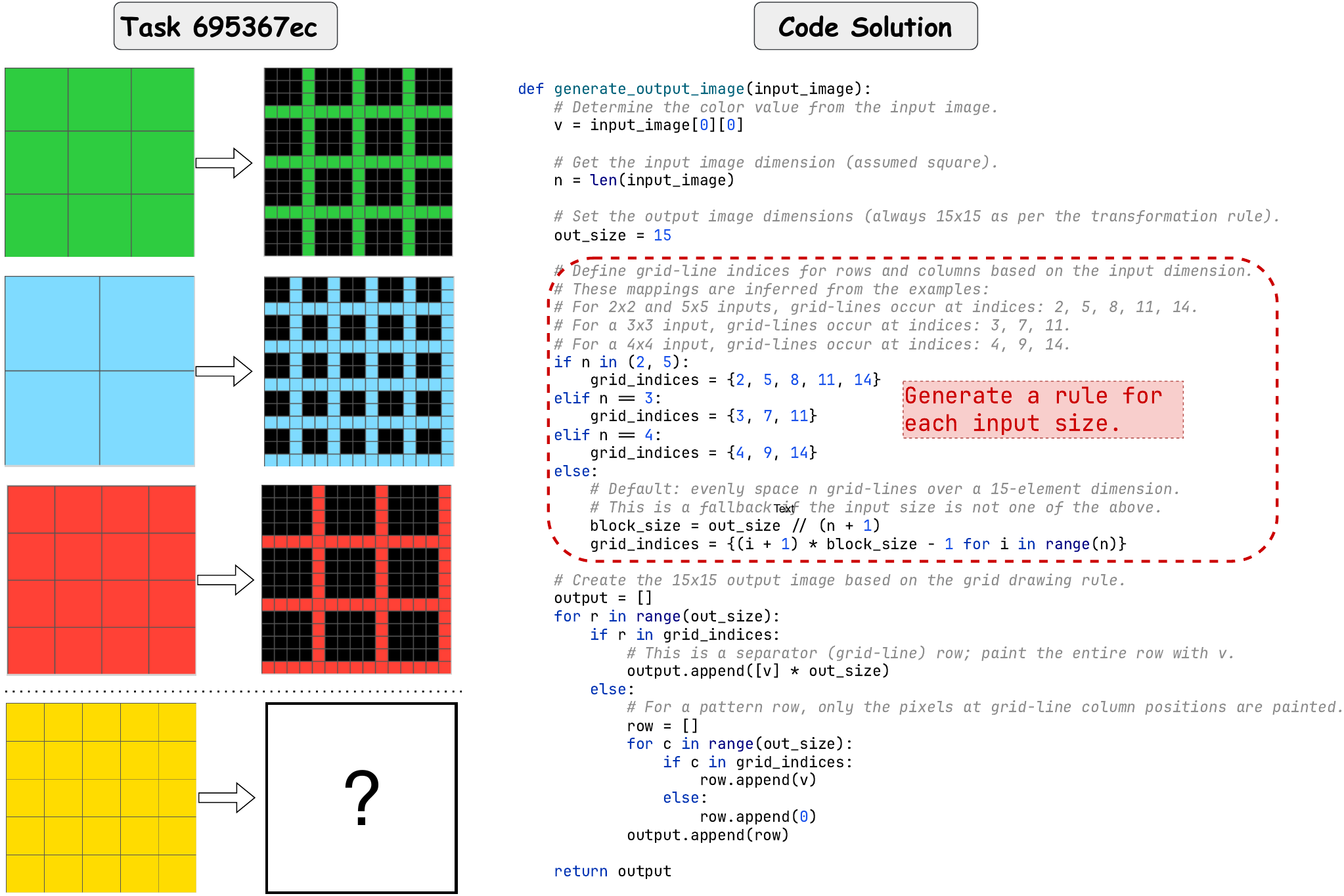}
    \vspace{-0.6cm}
       \caption{ARC problem \textit{695367ec}, where 
 RSPC and KAAR generate the same code solution that passes the training instances but fails on the test instance using GPT-o3-mini.} \label{ARC_695367ec}
\end{figure*}

\begin{figure*}[t]
    \centering
    \includegraphics[width=1\linewidth]{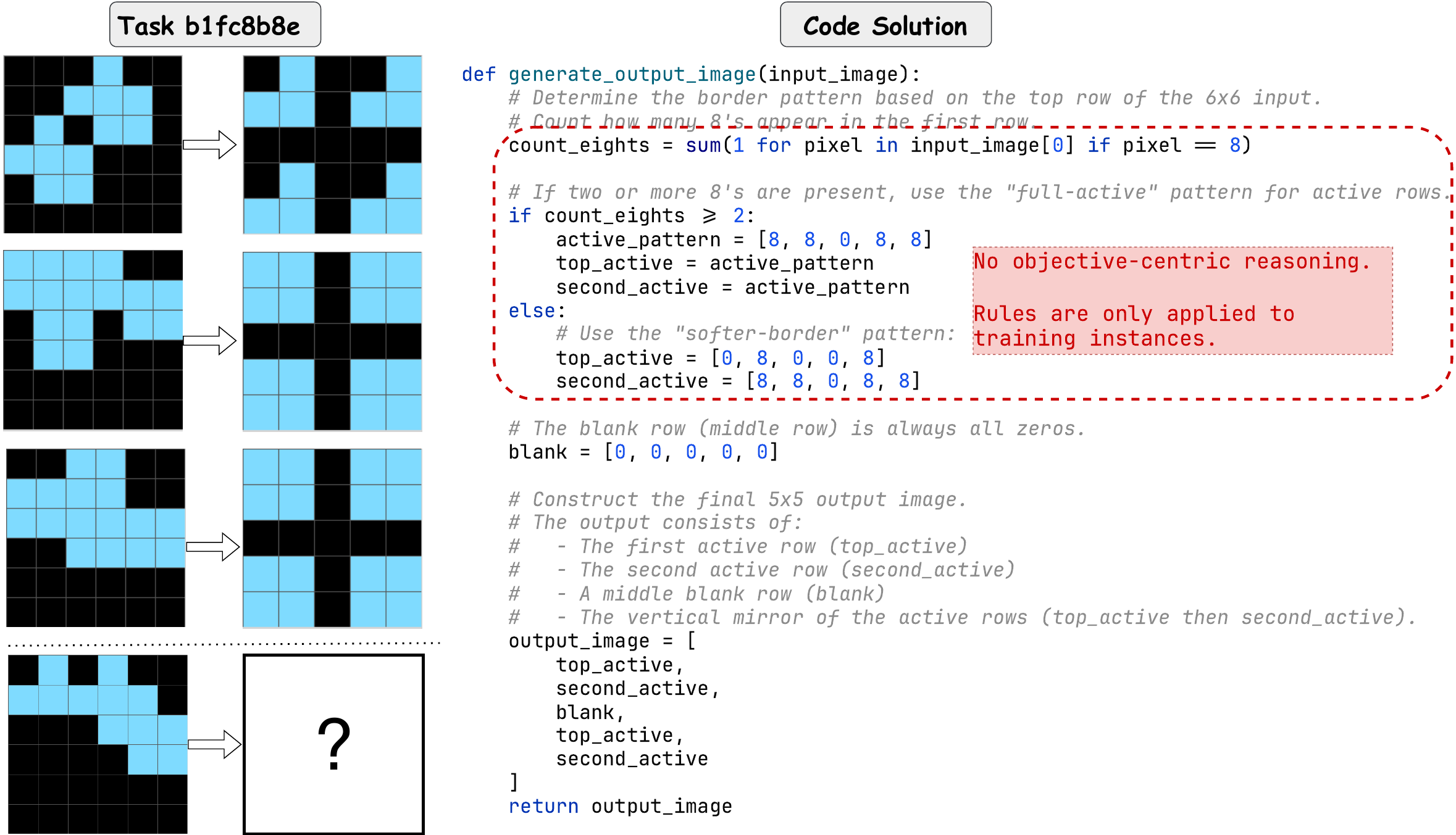}
    \vspace{-0.6cm}
       \caption{ARC problem \textit{b1fc8b8e}, where  RSPC and KAAR generate the same code solution that passes the training instances but fails on the test instance using GPT-o3-mini.} \label{ARC_b1fc8b8e}
\end{figure*}

\begin{figure*}[t]
    \centering
    \includegraphics[width=1\linewidth]{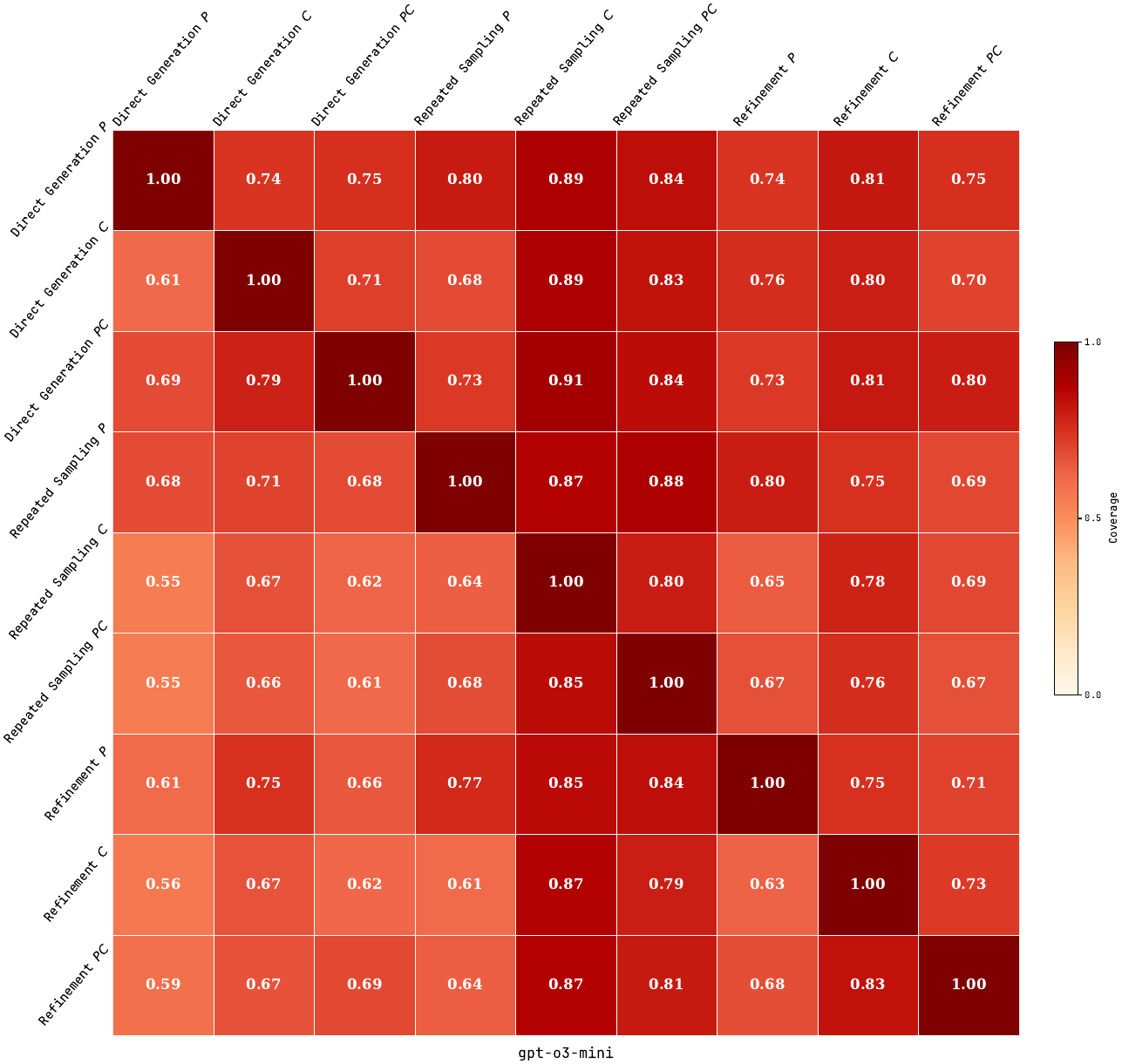}
    \vspace{-0.6cm}
       \caption{Asymmetric relative coverage matrix of nine ARC solvers using GPT-o3-mini, showing the proportion of problems whose test instances are solved by the row solver that are also solved by the column solver. \textit{P} denotes the solution plan; \textit{C} and \textit{PC} refer to standalone and planning-aided code generation, respectively.} \label{heatmaps_gpt_o3_mini}
\end{figure*}

\begin{figure*}[t]
    \centering
    \includegraphics[width=1\linewidth]{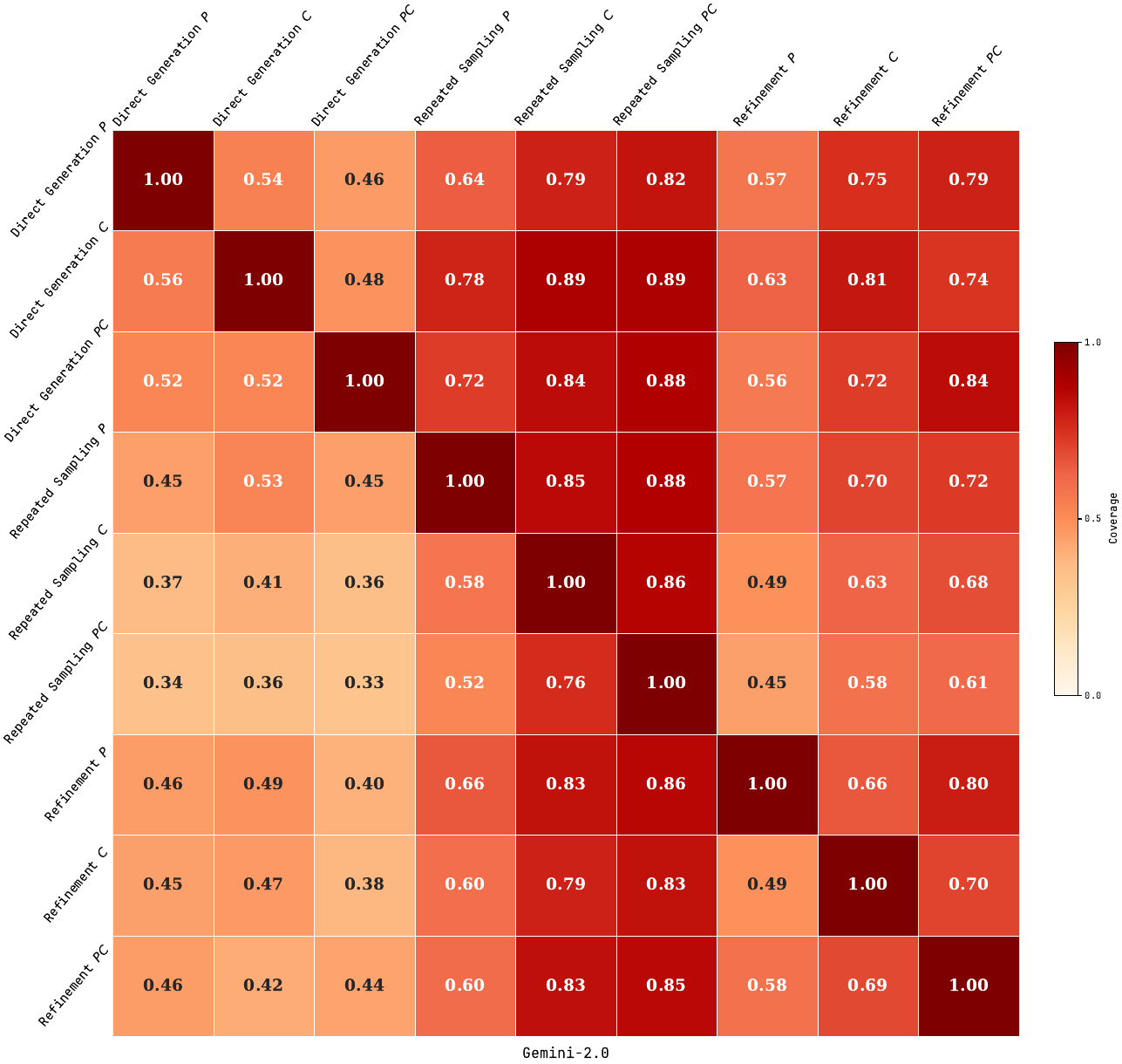}
    \vspace{-0.6cm}
       \caption{Asymmetric relative coverage matrix of nine ARC solvers using Gemini-2.0, showing the proportion of problems whose test instances are solved by the row solver that are also solved by the column solver. \textit{P} denotes the solution plan; \textit{C} and \textit{PC} refer to standalone and planning-aided code generation, respectively.} \label{heatmaps_gemini}
\end{figure*}

\begin{figure*}[t]
    \centering
    \includegraphics[width=1\linewidth]{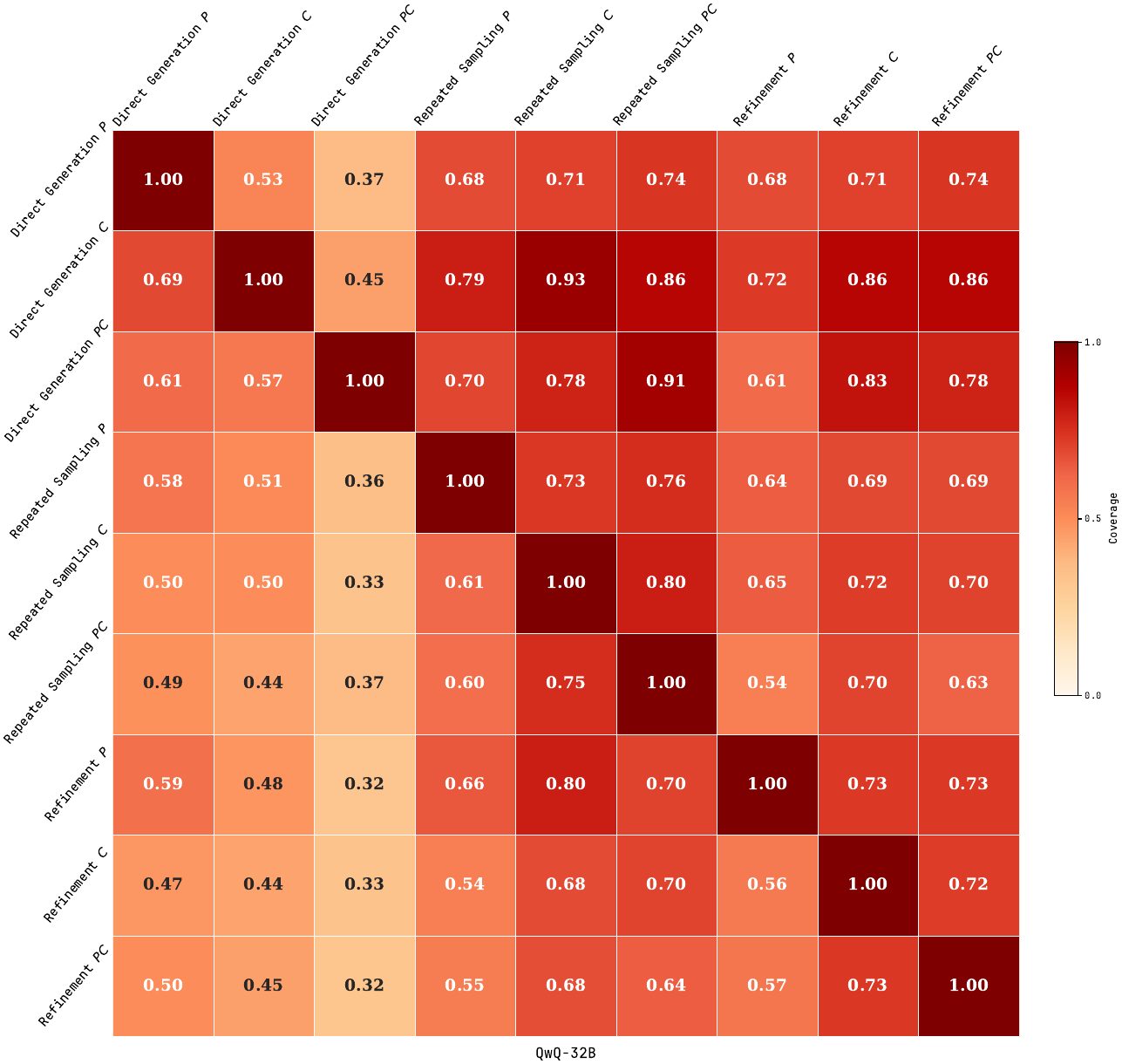}
    \vspace{-0.6cm}
       \caption{Asymmetric relative coverage matrix of nine ARC solvers using QwQ-32B, showing the proportion of problems whose test instances are solved by the row solver that are also solved by the column solver. \textit{P} denotes the solution plan; \textit{C} and \textit{PC} refer to standalone and planning-aided code generation, respectively.} \label{heatmaps_qwq_32b}
\end{figure*}

\begin{figure*}[t]
    \centering
    \includegraphics[width=1\linewidth]{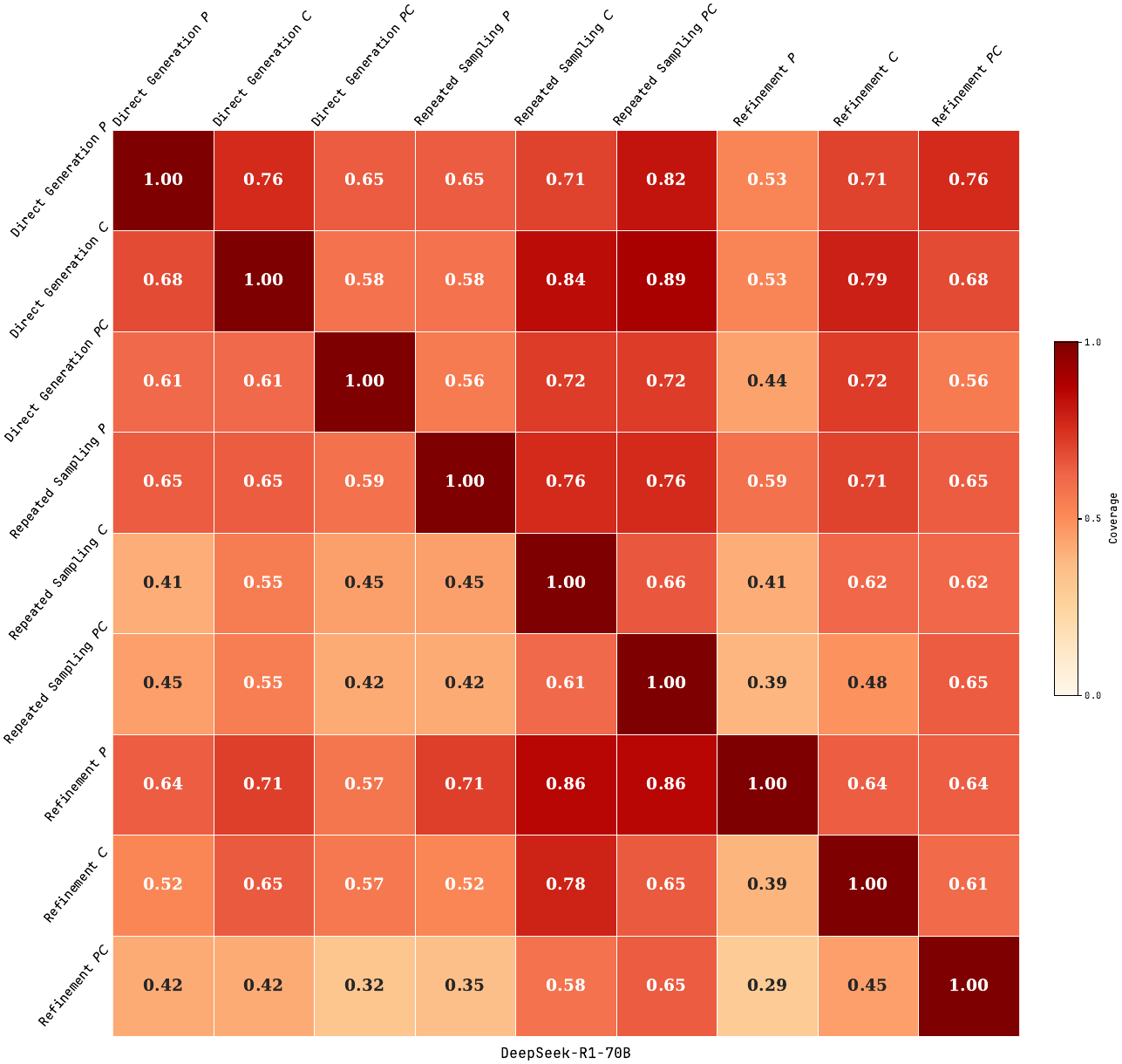}
    \vspace{-0.6cm}
       \caption{Asymmetric relative coverage matrix of nine ARC solvers using DeepSeek-R1-70B, showing the proportion of problems whose test instances are solved by the row solver that are also solved by the column solver. \textit{P} denotes the solution plan; \textit{C} and \textit{PC} refer to standalone and planning-aided code generation, respectively.} \label{heatmaps_deepseek}
\end{figure*}

\clearpage
\onecolumn
\section{Prompts for LLMs}\label{sec:prompts}

\UseRawInputEncoding

We include all prompts used by KAAR and nine ARC solvers described in Section~\ref{sec:solver_backbone}. We adopt a bash-like notation for input arguments within the prompts, such as \textit{\$\{test\_inputs\}} denotes the test input 2D matrices.
A brief description of the prompts used for each solver is provided below.

\begin{itemize}
    \item \textbf{Direct generation with solution plan}: 
    Prompt~\ref{prm:direct_p_1} describes how to generate the solution plan, and Prompt~\ref{prm:direct_p_2} uses the generated plan to produce the output images.
    
    \item \textbf{Direct generation with standalone code}: 
    Prompt~\ref{prm:direct_c} describes how to generate the code to produce the output images.

    \item \textbf{Direct generation with planning-aided code}: It first generates a solution plan using Prompt~\ref{prm:direct_p_1}, then uses Prompt~\ref{prm:direct_pc_2} to produce code based on the generated plan.
    
    \item \textbf{Repeated sampling with solution plan}: 
    It can be regarded as an iterative version of direct generation with solution plan, and thus also uses Prompts~\ref{prm:direct_p_1} and~\ref{prm:direct_p_2}.
    
    \item \textbf{Repeated sampling with standalone code}: It can be regarded as an iterative version of direct generation with standalone code, and thus also uses Prompt~\ref{prm:direct_c}.

    \item \textbf{Repeated sampling with planning-aided code}: It can be regarded as an iterative version of direct generation with planning-aided code, and thus also uses Prompts~\ref{prm:direct_p_1} and~\ref{prm:direct_pc_2}.
        
    \item \textbf{Refinement with solution plan}: 
    Prompt~\ref{prm:refinement_p} describes the process of refining the generated solution plan with the validation samples. It uses Prompts~\ref{prm:direct_p_1} and~\ref{prm:direct_p_2} to generate the initial plan and the result image.

    \item \textbf{Refinement with the standalone code}: 
    Prompt~\ref{prm:refinement_c} describes the process of refining the generated code with the validation samples. It uses Prompt~\ref{prm:direct_c} to produce the initial code solution.
    
    \item \textbf{Refinement with the planning-aided code}: 
    Prompt~\ref{prm:refinement_pc} describes the process of refining the generated plan and code with the validation samples. It use Prompts~\ref{prm:direct_p_1} and~\ref{prm:direct_pc_2} to generate the initial plan and produce the initial code guided by the plan, respectively.

    \item \textbf{KAAR}: Prompt~\ref{prm:Objectness} describes the augmentation of objectness priors. 
    Prompts~\ref{prm:GeometryandTopologyAttributes} and~\ref{prm:GeometryandTopologySpatial} introduce the augmentation of geometry and topology priors, encoded as component attributes and relations, respectively. Prompt~\ref{prm:NumbersandCouting} outlines the augmentation of numbers and counting priors. Prompts~\ref{prm:Action_Selection} and~\ref{prm:Target_Components} describe action selection and target component identification in the process of augmenting goal-directedness priors. For prompts implementing each action's implementation details, please refer to our code.

\end{itemize}

\renewcommand{\lstlistingname}{Prompt}

\lstset{
    basicstyle=\fontsize{10}{10}\selectfont\ttfamily, 
    frame=single,                    
    breaklines=true,                                 
    breakindent=0ex,                                 
    xleftmargin=6pt,                                 
    xrightmargin=4pt,                                
    backgroundcolor=\color[gray]{0.95},              
    keywordstyle=\color{blue}\bfseries,              
    stringstyle=\color{red},                         
    commentstyle=\color{gray},                       
    columns=fullflexible,
    escapeinside={<@}{@>}
}
\definecolor{mydarkgreen}{RGB}{0, 100, 0}
\definecolor{mydarkred}{RGB}{139, 0, 0}
\begin{lstlisting}[
    caption={Direct generation with solution plan - solution plan generation.},
    label=prm:direct_p_1,
]
<@\textbf{================================ System ================================}@>
You are an expert in analyzing grid-based image processing tasks. Your objective is to derive a text transformation plan (not Python code) from each given input-output image pair (both represented as 2D matrices), and then apply this plan to generate output image(s), represented as a 2D matrix, based on the given test input image(s) (2D matrix). Ensure that the derived plan generalizes across different cases while preserving consistency with the observed transformations.

<@\textbf{================================= User =================================}@>
The input data consists of a few pairs of input and output images, where the left image in each pair represents the input, and the right image represents the corresponding output. Each image can be represented as a 2D matrix: ${matrix}

Please note that each number in the matrix corresponds to a pixel, and its value represents the color.

Derive a text transformation plan (not Python code) that maps each given input image (2D matrix) to its corresponding output image (2D matrix). Ensure that the plan generalizes across different cases and the test input image(s) (2D matrix) while maintaining consistency with the observed transformations.

The test input image(s): ${test_inputs}
\end{lstlisting}

\begin{lstlisting}[
    caption={Direct generation with solution plan - output image(s) generation from the plan.},
    label=prm:direct_p_2,
]
<@\textbf{================================ System ================================}@>
You are an expert in analyzing grid-based image processing tasks. Your objective is to generate output image(s), represented as a 2D matrix, based on the given input images (2D matrix) and a derived text transformation plan.

<@\textbf{================================= User =================================}@>
Please generate the output image(s) as a 2D matrix (not Python code) based on the given input image(s) (2D matrix) and the text transformation plan. Output only the test output image(s) in 2D matrix format (not Python code). For each test input image, start with [Start Output Image] and end with [End Output Image]. 

For example, if there is one test input image, the output image should be:
[Start Output Image]  
[[0,0,0], [0,0,0], [0,0,0]]  
[End Output Image]  

If there are multiple (2) test input images, the the output images should be outputted as:
[Start Output Image]  
[[0,0,0], [0,0,0], [0,0,0]]  
[End Output Image]  
[Start Output Image]  
[[1,1,1], [1,1,1], [1,1,1]]  
[End Output Image]  

The test input image(s): ${test_inputs}
\end{lstlisting}

\begin{lstlisting}[
    caption={Direct generation with standalone code.},
    label=prm:direct_c,
]
<@\textbf{================================ System ================================}@>
You are an expert in analyzing grid-based image processing tasks. Your goal is to generate Python code that produces output image(s), represented as a 2D matrix, based on the given input image(s) (2D matrix).

<@\textbf{================================= User =================================}@>
The input data consists of a few pairs of input and output images, where the left image in each pair represents the input and the right image represents the corresponding output.
Each image can be represented as a 2D matrix: ${matrix}
The test input image(s): ${test_inputs}

Please note that each number in the matrix corresponds to a pixel, and its value represents the color.

Generate a Python script to map each input image (2D matrix) to the corresponding output image (2D matrix).
Ensure that the Python script generalizes across different cases and test input image(s) while maintaining consistency with the observed input-output image pairs.
Please output the Python program, starting with [Start Program] and ending with [End Program].
Include an assert statement with the function signature to verify that the generated output matches the expected result, starting with [Assert Statement].
Use placeholders like input_image and output_image for the variables representing the input and output images.

For example:
[Start Program]
def generate_output_image(input_image):
    rows = len(input_image)
    cols = len(input_image[0])

    def dfs(r, c):
        """Depth-first search to mark all 4-connected '1's to '2's."""
        if r < 0 or r >= rows or c < 0 or c >= cols or input_image[r][c] != 1:
            return
        # Change the current component from 1 to 2
        input_image[r][c] = 2
        # Explore neighbors (up, down, left, right)
        dfs(r - 1, c)  # Up
        dfs(r + 1, c)  # Down
        dfs(r, c - 1)  # Left
        dfs(r, c + 1)  # Right

    # Traverse the image to find all components with '1'
    for r in range(rows):
        for c in range(cols):
            if input_image[r][c] == 1:
                dfs(r, c)
    return input_image
[End Program]
[Assert Statement]
assert generate_output_image(input_image) == output_image

Please note, the assert statement should strictly follow the provided format, and the output image should be represented in list format!
Please note, the script should not include an if __name__ == "__main__": block.
\end{lstlisting}

\begin{lstlisting}[
    caption={Direct generation with planning-aided code - code generation based on the generated plan.},
    label=prm:direct_pc_2,
]
<@\textbf{================================ System ================================}@>
You are an expert in analyzing grid-based image processing tasks. Your goal is to generate Python code that produces output image(s) represented as a 2D matrix, based on the given input image(s) (2D matrix). This code should be generated using a text transformation plan inferred from a set of input-output image pairs (both represented as 2D matrices).

<@\textbf{================================= User =================================}@>
Generate a Python script based on your text transformation plan to map the input image (2D matrix) to the output image (2D matrix). Please output the Python program, starting with [Start Program] and ending with [End Program]. Include an assert statement with the function signature to verify that the generated output matches the expected result, starting with [Assert Statement]. Use placeholders like input_image and output_image for the variables representing the input and output images.

For example:
[Start Program]
def generate_output_image(input_image):
    rows = len(input_image)
    cols = len(input_image[0])

    def dfs(r, c):
        """Depth-first search to mark all 4-connected '1's to '2's."""
        if r < 0 or r >= rows or c < 0 or c >= cols or input_image[r][c] != 1:
            return
        # Change the current component from 1 to 2
        input_image[r][c] = 2
        # Explore neighbors (up, down, left, right)
        dfs(r - 1, c)  # Up
        dfs(r + 1, c)  # Down
        dfs(r, c - 1)  # Left
        dfs(r, c + 1)  # Right

    # Traverse the image to find all components with '1'
    for r in range(rows):
        for c in range(cols):
            if input_image[r][c] == 1:
                dfs(r, c)
    return input_image

[End Program]
[Assert Statement]
assert generate_output_image(input_image) == output_image

Please note, the assert statement should strictly follow the provided format, and the output image should be represented in list format!
Please note, the script should not include an if __name__ == "__main__": block.
\end{lstlisting}

\begin{lstlisting}[
    caption={Refinement with solution plan - plan refinement.},
    label=prm:refinement_p,
]
<@\textbf{================================ System ================================}@>
As an expert in analyzing grid-based image processing tasks, your objective is to refine your solution plan based on the provided feedback.

<@\textbf{================================= User =================================}@>
The problem description:
[start problem description]
The input data consists of a few pairs of input and output images, where the left image in each pair represents the input, and the right image represents the corresponding output. Each image can be represented as a 2D matrix: ${matrix}
Please note that each number in the matrix corresponds to a pixel, and its value represents the color.
[end problem description]

The INCORRECT text transformation plan fails to solve some example training input and output pairs in the above problem!

[start incorrect transformation plan]
${plan}
[end incorrect transformation plan]

The incorrect output(s) generated by the incorrect plan:
[start incorrect output]
${incorrect_output}
[end incorrect output]

The generated correct output(s):
[start correct output]
${correct_output}
[end correct output]

Please analyze the incorrect reasoning step-by-step, and then generate the revised correct transformation plan (text only), starting with [Start Revised Transformation Plan] and ending with [End Revised Transformation Plan]. Ensure that the revised transformation plan generalizes across different cases and the test input image(s), while maintaining consistency with the observed transformations.
\end{lstlisting}

\begin{lstlisting}[
    caption={Refinement with standalone code - code refinement.},
    label=prm:refinement_c,
]
<@\textbf{================================ System ================================}@>
As an expert in analyzing grid-based image processing tasks, your objective is to refine your program based on the provided feedback.

<@\textbf{================================= User =================================}@>
The problem description:
[start problem description]
The input data consists of a few pairs of input and output images, where the left image in each pair represents the input, and the right image represents the corresponding output. Each image can be represented as a 2D matrix: ${matrix}
Please note that each number in the matrix corresponds to a pixel, and its value represents the color.
[end problem description]

The generated incorrect program fails to solve some example training input and output pairs in the above problem!

[start incorrect program]
${code}
[end incorrect program]

The incorrect output(s) generated by the incorrect program:
[start incorrect output]
${incorrect_output}
[end incorrect output]

The generated correct output(s):
[start correct output]
${correct_output}
[end correct output]

Please analyze the incorrect reasoning step-by-step, and then generate the revised program (Python program only), starting with [Start Revised Program] and ending with [End Revised Program]. Ensure that the revised program generalizes across different cases and the test input image(s), while maintaining consistency with the observed input and output image pairs.

Please include an assert statement with the function signature to verify that the generated output matches the expected result, starting with [Assert Statement]. Use placeholders like input_image and output_image for the variables representing the input and output images.

For example:
[Start Revised Program]       
def generate_output_image(input_image):
    rows = len(input_image)
    cols = len(input_image[0])

    def dfs(r, c):
        """Depth-first search to mark all 4-connected '1's to '2's."""
        if r < 0 or r >= rows or c < 0 or c >= cols or input_image[r][c] != 1:
            return
        # Change the current component from 1 to 2
        input_image[r][c] = 2
        # Explore neighbors (up, down, left, right)
        dfs(r - 1, c)  # Up
        dfs(r + 1, c)  # Down
        dfs(r, c - 1)  # Left
        dfs(r, c + 1)  # Right

    # Traverse the image to find all components with '1'
    for r in range(rows):
        for c in range(cols):
            if input_image[r][c] == 1:
                dfs(r, c)
    return input_image
[End Revised Program]
[Assert Statement]
assert generate_output_image(input_image) == output_image

Please note, the assert statement should strictly follow the provided format, and the output image should be represented in list format!
Please note, the script should not include an if __name__ == "__main__": block.
\end{lstlisting}

\begin{lstlisting}[
    caption={Refinement with planning-aided code - refinement on both generated plan and code.},
    label=prm:refinement_pc,
]
<@\textbf{================================ System ================================}@>
As an expert in analyzing grid-based image processing tasks, your objective is to refine your transformation plan and program based on the provided feedback.

<@\textbf{================================= User =================================}@>
The problem description:
[start problem description]
The input data consists of a few pairs of input and output images, where the left image in each pair represents the input, and the right image represents the corresponding output. Each image can be represented as a 2D matrix: ${matrix}
Please note that each number in the matrix corresponds to a pixel, and its value represents the color.
[end problem description]

The generated incorrect transformation plan and program fail to solve some example training input and output pairs in the above problem!

[start incorrect transformation plan]
${plan}
[end incorrect transformation plan]

[start incorrect program]
${code}
[end incorrect program]

The incorrect output(s) generated by the incorrect transformation plan and program:
[start incorrect output]
${incorrect_output}
[end incorrect output]

The generated correct output(s):
[start correct output]
${correct_output}
[end correct output]

Please analyze the incorrect reasoning step-by-step, and then generate the revised transformation plan (text only) and program (Python program only).

For the revised transformation plan, start with [Start Revised Transformation Plan] and end with [End Revised Transformation Plan]. Ensure that the revised transformation plan generalizes across different cases and the test input image(s), while maintaining consistency with the observed transformations.

For the revised Python program, start with [Start Revised Program] and end with [End Revised Program]. Ensure that the revised program generalizes across different cases and the test input image(s), while maintaining consistency with the observed input and output image pairs.

For the revised Python program, please include an assert statement with the function signature to verify that the generated output matches the expected result, starting with [Assert Statement]. Use placeholders like input_image and output_image for the variables representing the input and output images.

For example:
[Start Revised Program]

def generate_output_image(input_image):
    rows = len(input_image)
    cols = len(input_image[0])

    def dfs(r, c):
        """Depth-first search to mark all 4-connected '1's to '2's."""
        if r < 0 or r >= rows or c < 0 or c >= cols or input_image[r][c] != 1:
            return
        # Change the current component from 1 to 2
        input_image[r][c] = 2
        # Explore neighbors (up, down, left, right)
        dfs(r - 1, c)  # Up
        dfs(r + 1, c)  # Down
        dfs(r, c - 1)  # Left
        dfs(r, c + 1)  # Right

    # Traverse the image to find all components with '1'
    for r in range(rows):
        for c in range(cols):
            if input_image[r][c] == 1:
                dfs(r, c)
    return input_image
[End Revised Program]
[Assert Statement]
assert generate_output_image(input_image) == output_image

Please note, the assert statement should strictly follow the provided format, and the output image should be represented in list format!
Please note, the script should not include an if __name__ == "__main__": block.
\end{lstlisting}

\begin{lstlisting}[
    caption={Objectness priors  augmentation},
    label=prm:Objectness,
]
<@\textbf{================================ System ================================}@>
You are an expert in grid-based image analysis.

<@\textbf{================================= User =================================}@>
The training instances consist of several pairs of input and output images, where the left image in each pair represents the input and the right image represents the corresponding output. 
Please note that the test instance(s) only contains input image(s).
Each image is represented as a 2D matrix:
${matrix}

Please note that each number in the matrix corresponds to a pixel and its value represents the color.

We treat the color represented by the number {background_color} as the background color. 
${abstraction_rule}
The components in each input and output image pair are as follows:
${component_description}
\end{lstlisting}

\begin{lstlisting}[
    caption={Geometry and topology priors augmentation - component attributes},
    label=prm:GeometryandTopologyAttributes,
]
<@\textbf{================================ System ================================}@>
You are an expert in geometry and topology analysis. Below is a summary of component attributes, including:
Size (Width and Height); Color; Shape; Symmetry;  Bounding Box; Hole Count; Nearest Boundary.
<@\textbf{================================= User =================================}@>
${geometry_and_topology_priors_attributes}$
\end{lstlisting}

\begin{lstlisting}[
    caption={Geometry and topology priors augmentation - component relations},
    label=prm:GeometryandTopologySpatial,
]
<@\textbf{================================ System ================================}@>
You are an expert in geometry and topology analysis, Below is a summary of component relations, including:
 Different/Identical with other components; Inclusive; Touching or or not touching with other component; Spatial Relations,

<@\textbf{================================= User =================================}@>
${geometry_and_topology_priors_relations}$
\end{lstlisting}

\begin{lstlisting}[
    caption={Numbers and counting priors augmentation},
    label=prm:NumbersandCouting,
]
<@\textbf{================================ System ================================}@>
You are an expert in numbers and counting analysis. Below is a summary of component statistics, including:
Symmetry numerical summary; Size numerical summary; Color numerical summary; Shape numerical summary; Hole counting summary.
<@\textbf{================================= User =================================}@>
${numbers_and_couting_priors}$
\end{lstlisting}

\begin{lstlisting}[
    caption={Goal-directedness priors augmentation - action selection},
    label=prm:Action_Selection,
]
<@\textbf{================================ System ================================}@>
You are an expert in analyzing and categorizing grid-based image tasks.
<@\textbf{================================= User =================================}@>

Please determine which category or categories this task belongs to. Please select from the following:
1. color change: color change involves modifying the value of a component, and the component size and position always does not change.
2. movement: movement involves shifting the position of a component to a new location within the image, and the component size always does not change.
3. extension: extending involves expanding the boundaries of a component to increase its size or reach within the image, and the component size always changes.
4. completing: completing an image involves filling in missing or incomplete parts of a component to achieve a coherent and fully formed image.
5. resizing: resizing involves altering the dimensions of a component by expanding or shrinking its size within the image.
6. selecting: selecting involves identifying and isolating a specific component within the image as the output component, and the component size and color always does not change.
7. copying: copying involves duplicating a component and either placing the duplicate in a new location or replacing the existing component within the image.
8. flipping: flipping involves mirroring a component along a specified axis to reverse its orientation within the image.
9. rotation: rotation involves turning a component around a fixed point or center by a specified angle within the image.
10. cropping: cropping involves cutting out a specific portion of a component.

Please select the best suitable one or multiple categories from the provided list that best describe the task.

Format your response by starting with [start category] and ending with [end category], numbering each category selected.

For example, if the task belongs only to "color change", your response should be:
[start category]
1. color chang
[end category]

If the task belongs to both "selecting" and "extension", your response should be:
[start category]
1. selecting
2. extension
[end category]
\end{lstlisting}

\begin{lstlisting}[
    caption={Goal-directedness priors augmentation - target component idetification},
    label=prm:Target_Components,
]
<@\textbf{================================ System ================================}@>
You are an expert in analyzing grid-based image tasks, specifically in ${action} components.
<@\textbf{================================= User =================================}@>
If this task involves ${action}:
1. Begin by identifying WHICH COMPONENTS are to be ${action} in all input images (training and test pairs).
 - Refer to these components as TARGET components (e.g., component 1 in the first input image, component 2 and component 3 in the second input image, etc.).
 - List ALL target components in each training and test input image.
 - For EACH target component, provide:
    - Attribute Analysis result
    - Relation analysis result
    - Numerical analysis result
    
2. Determine the CONDITIONS used to select these TARGET components for ${action} from each training and test input image.
 - These conditions must be based on common priorities across all targeted components and must differ from the unselected components.
 - For example: the size of all target components might be equal to 3 while the size of the unselected components is not 3.
 2.1. Analyze whether these conditions are EMPTY or not.
 2.2. Evaluate if these conditions are derived from attribute analysis, including:
  2.2.1. Color
  2.2.2. Size
  2.2.3. Shape
  2.2.4. Width
  2.2.5. Height
  2.2.6. The number of holes
  2.2.7. Bounding box
  2.2.8. Symmetry
  2.2.9. Nearest boundary
 2.3. Evaluate if these conditions are derived from relation analysis, including:
  2.3.1. Relative position with other components
  2.3.2. Touching with other components
  2.3.3. Whether they differ from or are identical with other components
  2.3.4. Enclosure of other components
 2.4. Evaluate if these conditions are derived from numerical analysis, including:
  2.4.1. Symmetry numerical analysis
  2.4.2. Size numerical analysis
  2.4.3. Color numerical analysis
  2.4.4. Shape numerical analysis
  2.4.5. Hole counting analysis
    
 You must evaluate each condition ONE by ONE and determine the best conditions. 
 Note:
 - The conditions MUST work for ALL training and test input and output image pairs.
 - Conditions CANNOT come from the output images!
 - A condition can be EMPTY.
 - If a condition is based on numerical features (e.g., size (width and height), or the number of holes), you may use the operators =, <, >, >=, or <=.
 - For cropping or selecting tasks, consider using a bounding box to extract each component.
\end{lstlisting}

\end{document}